\documentclass[preprint,nonacm]{acmart}

\AtBeginDocument{%
  \providecommand\BibTeX{{%
    \normalfont B\kern-0.5em{\scshape i\kern-0.25em b}\kern-0.8em\TeX}}}
\setcitestyle{authoryear, open={((}, close={))}}
\usepackage{subcaption}
\usepackage{libertine}
\usepackage{lipsum}% http://ctan.org/pkg/lipsum
\usepackage{algpseudocode}% http://ctan.org/pkg/algorithmicx
\usepackage{graphicx}
\usepackage{grffile}
\usepackage[compatibility=false]{caption}% http://ctan.org/pkg/caption
\usepackage{booktabs}
\usepackage{url}
\usepackage{multirow}
\usepackage{pgfplots}
\usepackage{tikz}
\usetikzlibrary{matrix,fit,shapes,calc,positioning,shadows,arrows,shapes,backgrounds,decorations.markings,fadings}
\usepgfplotslibrary{statistics}
\usepackage{listings}

\usepackage{balance}
\usepackage{wrapfig}
\usepackage{enumitem}
\usepackage[utf8]{inputenc}
\usepackage[T1]{fontenc} 
\usepackage[ruled,vlined]{algorithm2e} % Algorithm
\usepackage{color, colortbl}
\definecolor{Gray}{gray}{0.9}
\usepackage[skins]{tcolorbox}
\definecolor{bgBlock}{rgb}{0.22,0.15,0.49}
\definecolor{bgBlockAlert}{rgb}{0.99,0.84,0.31}
\definecolor{fgBlockAlert}{rgb}{0.22,0.15,0.49}
\definecolor{fgBlocxsk}{rgb}{0.99,0.84,0.31}
\definecolor{darkred}{rgb}{0.5,0,0}
\definecolor{darkgreen}{rgb}{0,0.5,0}
\definecolor{darkblue}{rgb}{0,0,0.5}
\definecolor{ForestGreen}{rgb}{0.13,0.54,0.13}
\definecolor{mypink3}{rgb}{0.15, 0.39, 0.04}

\usepackage{amsthm}

\theoremstyle{definition}

\newcommand{\ourcite}[1]{(\citet{#1})}

\newcommand{\tname}{\textsc{DAEGEN}}
\newcommand{\trepo}{\url{https://github.com/daegen/DaegenAttack}}
\newcommand{\trembarep}{\url{https://github.com/TransEmbedBA}}

\newcommand{\dx}{DeepXplore}
\newcommand{\df}{DLFuzz}
\newcommand{\simba}{Simba}
\newcommand{\tremba}{Tremba}
\newcommand{\inet}{ImageNet}

\newcommand{\ie}{i.e.}
\newcommand{\eg}{e.g.}
\newcommand{\etal}{et al.}
\newcommand{\nn}{$\mathit{NN}$}
\newcommand{\lnorm}{$L_2$}
\newcommand{\tm}[1]{$\mathit{#1}$}

\newcommand{\norm}[1]{$L_{#1}$}

\begin{document}
\title{Generating Adversarial Inputs Using A Black-box Differential Technique}
%\title{DAEGEN: A Black-box Differential Approach for Adversarial Input Generation}

\author{João Batista Pereira Matos Júnior}
\affiliation{%
\institution{Federal University of Amazonas}
\city{Manaus}
\state{Amazonas}
\country{Brazil}
}
\email{jbpmj@icomp.ufam.edu.br}
\author{Lucas Carvalho Cordeiro}
\affiliation{%
\institution{University of Manchester}
%\streetaddress{1 Th{\o}rv{\"a}ld Circle}
\city{Manchester}
\country{United Kingdom}}
\email{lucas.cordeiro@manchester.ac.uk}
\author{Marcelo d'Amorim}
\affiliation{%
\institution{Federal University of Pernambuco}
\city{Recife}
\state{Pernambuco}
\country{Brazil}
}
\email{damorim@cin.ufpe.br}

\author{Xiaowei Huang}
\affiliation{%
\institution{University of Liverpool}
\city{Liverpool}
%\state{Pernambuco}
\country{United Kingdom}
}
\email{xiaowei.huang@liverpool.ac.uk}

% \affiliation{%
% \affiliation{%
% \affiliation{%
% %% By default, the full list of authors will be used in the page
% %% headers. Often, this list is too long, and will overlap
% %% other information printed in the page headers. This command allows
% %% the author to define a more concise list
% %% of authors' names for this purpose.
%% The abstract is a summary of the work to be presented in the
%% articsreportle.
%  For example, \nn{}s can be used to classify the type of traffic signal from the image of that signal. 
\begin{abstract}
Neural Networks (\nn{}s) are known to be vulnerable to adversarial
attacks. A malicious agent initiates these attacks by perturbing an
input into another one such that the two inputs are classified
differently by the \nn{}.  In this paper, we consider a special class
of adversarial examples, which can exhibit not only the weakness of
\nn{} models---as do for the typical adversarial examples---but
also the different behavior between two \nn{} models. We call them
difference-inducing adversarial examples or DIAEs.  Specifically, we
propose \tname, the first \emph{black-box differential} technique for
adversarial input generation.  \tname\ takes as input two \nn{} models
of the same classification problem and reports on output an
adversarial example. The obtained adversarial example is a DIAE, so
that it represents a point-wise difference in the input space between
the two \nn{} models.  Algorithmically, \tname{} uses a local
search-based optimization algorithm to find DIAEs by iteratively
perturbing an input to maximize the difference of two models on
predicting the input. We conduct experiments on a spectrum of
benchmark datasets (\eg{}, MNIST, ImageNet, and Driving) and \nn{}
models (\eg{}, LeNet, ResNet, Dave, and VGG). Experimental results are
promising. First, we compare \tname{} with two existing white-box
differential techniques (DeepXplore and DLFuzz) and find that under
the same setting, \tname{} is 1)~\emph{effective}, i.e., it is the
only technique that succeeds in generating attacks in \emph{all}
cases, 2)~\emph{precise}, \ie{}, the adversarial attacks are very
likely to fool machines and humans, and 3)~\emph{efficient}, \ie{}, it
requires a reasonable number of classification queries. Second, we
compare \tname{} with state-of-the-art black-box adversarial attack
methods (\simba{} and \tremba{}), by adapting them to work on a differential setting. The experimental results show that \tname{} performs better than both of them. 
\end{abstract}
\maketitle
%--------------------------------------------------------
\section{{Introduction}}
\label{sec:introduction}
%--------------------------------------------------------
\sloppy In 2016, a Tesla vehicle, driving in autopilot mode, crashed
against a white truck after it failed to identify the truck as an
obstacle~\ourcite{Vlasic2018}. Further investigation revealed that the
autopilot system misclassified the truck as a bright sky. The Tesla
driver died in the crash. In 2018, a pedestrian was struck and killed
by an Uber test vehicle after it failed to recognize the obstacle as a
human~\ourcite{Lubben2018}. These are examples that illustrate the
importance of properly testing the software that controls these
vehicles.

Neural networks (\nn{}s) are computing systems capable of learning
tasks from examples. \nn{}s is the technology used to control the
vehicles mentioned above. They have been recently applied in various
domains, including image classification~\ourcite{Guo2017}, malware
detection~\ourcite{Yeo2018}, speech recognition~\ourcite{Juang2007},
medicine~\ourcite{SENGUPTA2016118}, and vehicle control and trajectory
prediction~\ourcite{Zissis2015}.  \nn{}s are known to be vulnerable to
\emph{adversarial
  attacks}~\ourcite{Szegedy2013IntriguingPO,Goodfellow2014,Papernot2017}. A
malicious agent creates these attacks by finding a pair of similar
inputs---typically, indistinguishable by the human eye---where the
\nn\ produces different outputs, one of which is incorrect. With that
knowledge, a malicious attacker can exploit neural-network-bearing
systems to perpetrate illicit acts. Adversarial attacks pose a
significant threat to the reliability of machine learning systems
today.  The Tesla and Uber accidents could have been prevented had the
autonomous vehicles been tested against adversarial attacks.  There exists
intensive research on adversarial attacks and defences today (See
\citet{DBLP:journals/corr/abs-1810-00069} and \citet{Huangsurvey2018}
for recent surveys). 

\emph{DIAEs.}  This paper focuses on a particular class of adversarial
examples that can differentiate two \nn{} models trained on the same
dataset. As~\citet{pei-etal-sosp2017} we use the term DIAE (for
Difference-Inducing Adversarial Examples) to refer to this class of
inputs.  The study of the differential behaviour between two \nn{}
models has been motivated by the current debate between the
transferability of adversarial examples~\ourcite{Papernot2017} and the
fact that decision boundaries of two models can be drastically
different even if they are trained on the same
dataset~\ourcite{LCLS2017}. Because the adversarial examples lie at
decision boundaries, the difference on decision boundaries suggests
the non-transferability of adversarial examples. In this paper,
instead of directly participating on this debate, we provide a
technique and corresponding tool to identify DIAEs, \ie{}, those
adversarial examples that perform differently on the two models. The
discovery of a DIAE suggests a point-wise failure of transferability
and a point-wise difference in the decision boundaries of two \nn{}
models.

\emph{Black-box Attacks.} White-box techniques exist to generate
DIAEs~\ourcite{pei-etal-sosp2017}. We are interested in black-box
methods for its practicality. Black-box methods rely only on the input
and outputs of the \nn{}s to produce adversarial examples. They do
\emph{not} need access to the gradients of a \nn, which is the key
ingredient used in white-box attacks, including the most successful
ones, such as C\&W attack~\ourcite{Carlini2016TowardsET}.
In simple words, we run both techniques on a single target \nn{}, 
saving the adversarial inputs they generate. 
Then, we later select a second \nn{} model and check if the outputs of this model match the outputs of the target model on these adversarial inputs. We consider the number of mismatches to be the number of difference inducing inputs.

\emph{Solution.}~This paper proposes \tname{}, a lightweight black-box
technique that leverages the diversity of different \nn{}s to generate
adversarial attacks. \tname{} is the first black-box technique to use
Differential Testing~\ourcite{McKeeman98differentialtesting} to find
DIAEs. For a given seed input $x$, \tname{} searches for adversarial
inputs $x'$ that maximize the distance between the outputs that the
two models produce to $x'$ (as to fool machines) while minimizing the
distance between $x$ and $x'$ (as to fool humans). We conjecture that
the combination of these goals enables \tname{} to produce
high-quality adversarial inputs efficiently. Although differential
testing has been used in testing of \nn{}s, specifically
DeepXplore~\ourcite{Pei2019} and DLFuzz \ourcite{guo-etal-fse2018},
both of them are white-box. In addition, there are significant
technical differences between \tname{} and these two methods. Unlike
\tname{} which finds DIAEs by minimising the difference between
predictive confidences of two \nn{} models, DeepXplore focuses on
neuron coverage---a metric of test adequacy--and is designed to find
DIAEs by ensuring that more neurons in one of the models are
activated. In other words, DeepXplore does not aim at exploring the
different behavior between \nn{} models--it explores the difference
between models to study adversarial behavior in one of the models. In
contrast, DLFuzz takes a single model on input instead of more than
one, as DeepXplore and \tname{} do. DLFuzz is not designed to work
with DIAEs.

\emph{Summary of Evaluation.}~We conducted an extensive set of
experiments to validate the performance of \tname{} by comparing it
with state-of-the-art white-box (differential) techniques and
black-box (non-differential) techniques. A simple adaptation (details explained in Section~\ref{sec:eval}) is needed to make those  non-differential techniques work on our differential setting.   For white-box techniques, we
compared \tname{} against \dx{}~\ourcite{Pei2019} and
\df{}~\ourcite{guo-etal-fse2018}. For black-box techniques, we
compared \tname{} against \simba{}~\ourcite{Guo2019Simba} and
\tremba{}~\ourcite{Huang2020}. We analyzed these techniques on three
orthogonal dimensions: 1)~\textit{effectiveness},
2)~\textit{precision}, and 3)~\textit{efficiency}. Effectiveness
refers to the number of times a technique was successful in generating
DIAEs. Precision refers to the quality of the adversarial inputs
generated. Intuitively, more similar inputs are preferable as they are
more difficult to catch by the human eye. Efficiency refers to the
computation cost of the technique. We used three popular image
datasets in the evaluation, namely, MNIST~\ourcite{Lecun1998},
\inet{}~\ourcite{Deng2009}, and Driving~\ourcite{udacity-challenge}
and several \nn{} models frequently used in the evaluation of related
techniques.

%% This looked a bit out of context to me -Marcelo
%% To the best of our knowledge, there exists no technique to date that works directly with differential behavior. In this paper, we adapt the state-of-the-art adversarial example generation techniques (Simba \cite{Guo2019Simba} and TREMBA \cite{Huang2020}). \Joao{Please review the next sentences: } We run both techniques as they are supposed to work on a single target \nn{}, saving the adversarial inputs they generate. Then, we later select a second \nn{} model and check if the outputs of this model match the outputs of the target model on its own adversarial inputs. We consider the number of mismatches to be the number of difference inducing inputs. Our experimental results show that \tname{} performs better than both of them in terms of difference inducing inputs generation.

\emph{Summary of Results.}~Considering the comparison with white-box
techniques, \tname{} was the only technique capable of generating
adversarial examples in \emph{all} cases and it was the fastest
technique. Considering the comparison with black-box techniques,
\tname{} was able to produce DIAEs in 99.2\% of the cases whereas
\simba\ and \tremba\ produced DIAEs on rates of 73.1\% and 76.9\%,
respectively. Furthermore, DAEGEN produced images with much lower
\lnorm values (\ie, Euclidean distance of the generated input to the
seed image) than \tremba\ and slightly higher \lnorm\ values than
\simba. Overall, these results provide initial, yet strong evidence
that our approach is effective to produce difference-inducing inputs
at higher rates.

%% \Mar{Revise this when we fix results --->}
%% \Fix{
%% \tname{} was the only technique to produce adversarial
%%   attacks in every case....  
%% \tname{} was superior to all black-box techniques on all metrics
%% analyzed. It outperformed the white-box differential technique on
%% effectiveness and efficiency. The precision of the white-box approach
%% was superior, as expected, but the difference was small.} Overall,
%% these results provide initial, yet strong evidence that our approach
%% is effective and differential testing can help improve black-box
%% approaches to adversarial input generation.
%-----------------------------------------------------------------
%\subsection{Contributions}
%-----------------------------------------------------------------
%We formalize the adversarial sample generation as an optimization problem and apply Hill Climbing and Genetic Algorithm 
%to randomly introduce small perturbations to an input image and search for a feasible solution, 
%therefore configuring successful black-box adversarial attacks.
This paper makes the following original contributions:
\newcommand{\Contrib}[1]{$\star$#1}
%[topsep=.2ex,itemsep=.2ex,leftmargin=1.1em]
\begin{itemize}[itemsep=.80ex]
  
\item[\Contrib{}] \textbf{Technique.}~\tname{}, the first black-box
  technique for finding DIAEs;
%  adversarial input  generation.
\item[\Contrib{}] \textbf{Tool.}~A publicly available tool implementing \tname;
\item[\Contrib{}] \textbf{Evaluation.}~A comprehensive evaluation of
  \tname{} including a variety of white- and black-box techniques,
  datasets, and \nn{} models. Experimental results indicate that \tname{} is
  effective, precise, and efficient.  Evaluation artifacts are
  publicly available at the following link \textbf{\trepo{}}.
  
  %% produces adversarial examples that are effective across
  %% different networks requiring a lower number of queries on MNIST and
  %% ImageNet when compared to other black-box attacks on the literature.
  %% \tname{} required 45-98\% fewer queries than state-of-art black-box
  %% threat methods to generate an adversarial example for MNIST images
  %% and about one hundred fewer queries for ImageNet images.
  %% Given the current knowledge in adversarial example generation, we believe it to be the first work to consider Hill Climbing to generate adversarial examples and promote adversarial attacks in a black-box setting.
 
\end{itemize}
\section{{Example}}
\label{sec:example}
%--------------------------------------------------------
Consider that a manufacturer of an autonomous vehicle wants to test
the robustness of the \nn{} models before they are deployed on their
vehicles. The focus is on a specific \nn{} that outputs a correction
angle on the steering wheel based on what it observes from the
pictures that the front camera of the vehicle takes. For example, the
\nn{} would instruct the vehicle to adjust the steering wheel's angle
to prevent the vehicle from falling off an approaching cliff. Other
nets on the vehicle could be used to monitor other environmental signals and to take other actions (\eg{}, whether or not to slow down or stop). This example focuses on \textit{driving direction}.

\tname{} is a technique to test the robustness of an \nn{}. It takes
as input a seed object (\eg{} an image) and two \nn{} models---the
testing target and the surrogate model for the target. As a black-box
technique, \tname\ uses information on the input and output of the net
to search for examples.  There are multiple ways to obtain surrogate
models. First, as shown by \citet{DBLP:journals/corr/PapernotMGJCS16},
a surrogate model can be created from scratch, by training over a set
of training data sampled from the model to be tested.  Second, in some
application domains, surrogate models may be available
off-the-shelf. For example, on benchmark datasets such as ImageNet,
there exist many well-trained models publicly available. The second
way is becoming a trend in self-driving applications, with more and
more benchmark datasets released (e.g., Waymo's Open
Dataset~\ourcite{waymo}).

%These surrogate models can may use network
%architecture and features.

\begin{figure}[t!]
  \centering
  \includegraphics[trim=0 0 0 0,clip,width=0.8\linewidth]{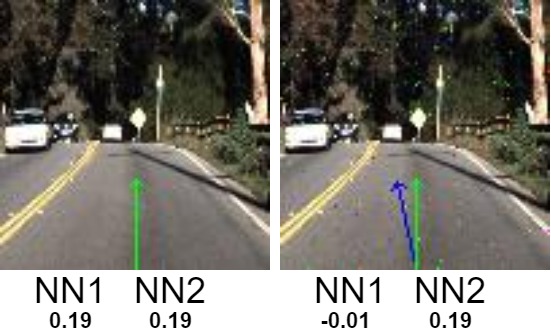}
  \caption{\label{fig:example}Original and adversarial input.}
  \vspace{-3ex}
\end{figure}

Figure~\ref{fig:example} shows an example of an adversarial input
produced by \tname{}. The seed (original) input to test the \nn{}
appears on the left-hand side of the figure. Note from the numbers
below the image on the left that both the tested net (NN1) and the
surrogate net (NN2) produce a similar response to this input---a .19
degree to the right (as the angle is positive). The angles are not
identical; we only showed two decimals in the figure for space. The
right side of the figure illustrates a very similar, but \emph{not}
identical, image for which the tested network and the surrogate
network differ, by a considerable amount, on the result they
produce. The network NN1 instructs the vehicle to go almost straight,
whereas the NN2 instructs the vehicle to steer to the
right. Regardless of which network is incorrect, there exists a
problem.

%, and the ensemble method has been widely considered.
We remark that our focus is not on the differential behavior of the
\nn{} models on the original input, as studied by
~\citet{Johansson2007, Abdullahi2015}. Instead, we study the different
behavior of the \nn{} models on the perturbed inputs, as shown in the
right-hand side of Figure~\ref{fig:example}. Our focus is on the
different robustness capability of the models and on identifying such
adversarial inputs. A DIAE shows a difference in the robustness
capability of two nets. Unlike general adversarial examples, whose
actual safety risk is arguable
\ourcite{2019arXiv190502175I,DBLP:journals/corr/LuSFF17}, DIAEs, which
are special kinds of adversarial examples, represent an actual risk
because they show a concrete manifestation of disagreeement between
two models.

\tname{} uses a classical Hill-Climbing search~\ourcite{Russell2009} to produce adversarial examples. It 1)~randomly selects a pixel from the image, 2)~makes random perturbations on that pixel, 3)~queries the models and measures \emph{fitness} of the new image, and 4)~stops if a sufficiently good adversarial input (according to fitness value) has been produced or it reached a budget on the number of queries to the models. Otherwise, if there was no fitness improvement, \tname{} continues execution on step 1 with the image it initiated the current iteration. If there was fitness improvement, it continues execution on step 1 with the image modified in the current iteration.

The sensible part of \tname{} is the objective function it uses to
search for examples. As a black-box technique, \tname{} only uses
information available on the input and outputs. The objective function
expresses two primary goals: 1)~to maximize the distance between the outputs
produced by the test and the surrogate networks (0.21 in the
adversarial example above) and 2)~to minimize the distance between the
inputs. \tname{} uses the Euclidean distance between two images to
that end. The rationale is that these two goals reflect the intention
of an adversarial attack---to fool machines (as per the first goal)
and humans (as per the second goal).

%\Mar{->}\Fix{Joao, please, report **numbers** of other techniques on this
%    example and a high-level justification explaining why DAEGEN works
%    and other don't.}

%\xiaowei{it might be helpful to draw an illustrative diagram to explain the effectiveness of differential testing in this context. Basically, two models may have different gradient directions and different decision boundaries. That is, along the gradient direction of a model will decrease the confidence of one model but may lead to the increase on the other model.  Also, the two models may have different decision boundaries. }

%--------------------------------------------------------
\section{{Preliminaries}}
\label{sec:preliminaries}
%--------------------------------------------------------

%Methods for adversarial attacks, which consist of systematically applying adversarial input generation approaches to exploit NNs systems, are usually classified based on two aspects: a) the attack goal, and b) the attacker capabilities~\ourcite{Serban2018AdversarialE}~\ourcite{Yuan2017AdversarialEA}. Regarding the attack goal, the possibilities of what the attacker can do are restricted to three domains: \textit{confidence reduction} \textcolor{red}{add citation}, \textit{random misclassification} \textcolor{red}{add citation}, and \textit{targeted misclassification} \textcolor{red}{add citation}.

This section introduces fundamental concepts and terminology used in
the paper.

% They are typically classified according to: a) the goal of the
% attack, and b) the capabilities of the
% attack.

Adversarial attacks apply input generation methods to exploit
NNs~\ourcite{Serban2018AdversarialE, Yuan2017AdversarialEA}.
Given an input $x$, the attacker goal is to look for another input
$x'$ such that one of the following objectives is satisfied:
confidence reduction (i.e., the confidence of classifying $x'$ as a
given label is reduced to a certain level), non-targeted
misclassification (i.e., $x'$ is classified into any other label that
is different from $x$), and targeted misclassification (i.e., $x'$ is
classified into a given label that is different from
$x$). Non-targeted and targeted attacks are specialized forms of
confidence reduction attacks. \emph{This paper focuses on non-targeted
  attacks}.

Attack methods differ in how much access they have on the \nn. Attacks
can be \textit{white-box} or \textit{black-box}.  A white-box attacker
has unlimited access to the \nn{}. For example, the attacker knows the
hyper-parameters, architecture, model weights, data used to train the
network model, and defense methods.  Hence, the attacker can replicate
the model under attack. In particular, a white-box attacker can
compute the gradient of loss for an input. The FGSM attack
\ourcite{Szegedy2013IntriguingPO}, for instance, leverages that
information. In contrast, black-box attacks have limited access to the
\nn{}. For a given input $x$, the attacker will only know the
distribution of the output or partial information about the output
distribution, as discussed below. \textit{This paper proposes a
  black-box attack method}.

%\theoremstyle{definition}
%\begin{definition}{(\textit{Black-box attack})}
%A black-box attack is characterized by the fact that the attacker is expected to have limited access to knowledge about the \nn{}. Except for the output of the network, any of the information available in white/gray box attacks are unavailable in black-box attacks. In particular, the \nn{} is treated as a black-box function, meaning that, for a given input $X$, the attacker will only know that the network produces an output of $y$. 
%\end{definition}

%Similar to~\ourcite{Ilyas2018, Chen2017Zoo}, we assume that the ability to access a \nn{} model means the ability to query a model, that is, providing an input image and retrieving the \nn{} output.
 
While white-box attack are appealing and many attack methods use the
gradients of a network, black-box attacks have a practical
edge. Existing literature describes three characteristics of black-box
attacks \ourcite{Ilyas2018}. The first one is limited query, where the
attacker may, for example, be limited to a small number of
queries. Many commercial or proprietary systems may limit the ability
of the attacker to access its classifiers. This characteristic
highlights the importance of query efficiency in black-box attacks,
thereby setting the ground for using the number of queries as a metric
of performance of the attack methods (see
Section~\ref{sec:metrics}). The second characteristic is partial
information. Attackers may have access to only the top-$k$ class
probabilities when querying the model, instead of the probability
distribution of all classes. \textit{\tname{} works with partial
  information; it only uses the top class from the probability
  distribution}. Finally, the third characteristic is label only,
where the attackers cannot access the probability
distributions. Instead, the only information available is the top-$k$
inferred labels. \textit{\tname{} does not work with this
  characteristic as it needs both the top-1 label and its
  probability}.

%------------------------------------------------------------------
\subsection{Black-box attack formulation}
%------------------------------------------------------------------
%Consider  the  existence  of  a  hypothetical system S whose purpose is to classify RGB images relevant in a social domain of human activity. Consider also that the classification is done by a NN model M which is not publicly available. We know that 

%The goal is to perform an adversarial attack, i.e., use an attack
%method \tm{A}, that applies a function \tm{Z} to generate and add
%perturbations on input image \tm{X}. By adding perturbations to the
%input image X, A generates an adversarial input image
%$X^\prime$. Considering that \tm{y} is the original label of \tm{X},
%and that the system \tm{S} classifies $X^\prime$ into some other
%label $y^\prime$ such that $y^\prime \neq y$, the attack is then
%successful.

We formally describe a black-box adversarial attack in the following.
Consider that a \nn{} model $S$ is trained to classify images into a
set of $k$ classes (\ie{}, labels).  \tm{S} can be formalized as a
function:
\[ f : {\Bbb R} ^d\rightarrow [0, 1]^k,\]

\noindent where \tm{d} is the number of input features.
The output of function \tm{f} is a probability distribution,
representing the probability of \tm{S} classifying the input \tm{x} into any of the \tm{k} labels. 
Formally, when \tm{S} is queried with the input image \tm{x}, it
responds by returning a tuple containing, for example, label $y \in 0...k$
and a probability $P(y|x) \in 0..1$. Non-targeted misclassification
can be formulated as an optimization problem~\ourcite{Pei2019}, as follows:
\begin{equation}
 \centering
 \label{equ:max}
 \min_{x' \in {\Bbb R} ^d} \|x - x^{\prime}\|_2 \text{~s.t.}  \arg\max_{t \in 1..k} f(x) \neq  \arg\max_{t \in 1..k} f(x^\prime)
\end{equation}
%\begin{equation}
% \centering
% \label{equ:max}
% \max_{x' \in {\Bbb R} ^d,~ t \in 1..k} \|f(x)_t - f(x^\prime)_t\| \text{, such that, } \|x - x^{\prime}\|_2 \leq \epsilon
%\end{equation}

The goal is to find an adversarial image \tm{x'}, where the difference between $x$ and $x^\prime$ is minimized when the label changes. 
Perturbation norms are used to define the acceptable distance between original and perturbed images~\ourcite{Bhambri2019ASO}. 
Specifically, the expression $\|x - x^{\prime}\|_2$ denotes the
\lnorm\ (Euclidean) distance between the inputs \tm{x} and \tm{x'}. In
addition to \norm{2}, \norm{1} and \norm{\infty} are also commonly
used. The right side of the equation expresses the condition that the
label needs to change for the attack to have an effect.

\section{A Black-box Differential Approach for Adversarial Input Generation (\tname{})}\label{sec:daegen}
\label{subsec:climat}
%---------------------------------------------

\begin{figure*}[t!]
  \vspace{-2ex}
  \centering
  \includegraphics[width=\linewidth]{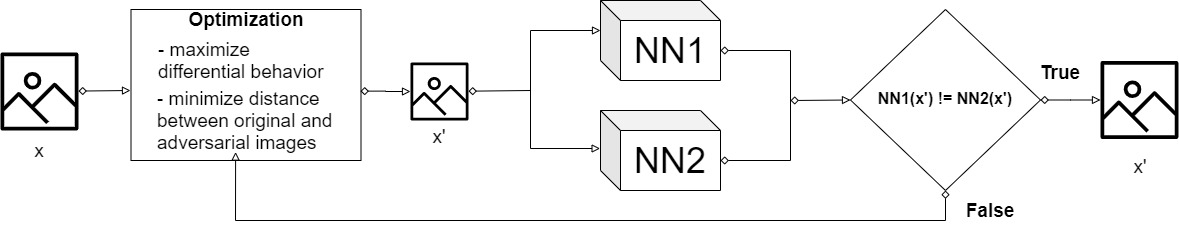}
  \vspace{-2ex}    
  \caption{Overview of the \tname{}`s architecture.}
  \vspace{-2ex}  
  \label{fig:DAEGEN}
\end{figure*}

This section describes \tname{}, a black-box differential method to find adversarial inputs in \nn{}s. Figure~\ref{fig:DAEGEN} summarizes the workflow of \tname{}. \tname{} uses search to identify adversarial inputs in \nn{}s.  However, in contrast to state-of-the-art white-box (differential) techniques, e.g., \ourcite{Pei2019} and \ourcite{guo-etal-fse2018} and black-box (non-differential) techniques, e.g., \ourcite{Guo2019Simba} and \ourcite{Huang2020}, \tname{} is inspired by the recent success of \textit{fuzzing} \ourcite{Godefroid20}, which is a highly effective, mostly automated, security testing technique. In particular, in evolutionary black-box fuzzing, if a mutation of the input triggers a new execution path through the code, then it is an exciting mutation; otherwise, the mutation is discarded. By producing random mutations of the input and observing their effect on code coverage, evolutionary black-box fuzzing can learn what interesting inputs are. Based on this observation, \tname{} queries the \nn{}s with given input and then makes perturbations (mutations) on the input based on observations obtained from the previous queries (learning). This process is repeated until finding an adversarial input or reaching a budget on the number of queries. 

%% In this section, we describe \tname{}, which is 
%% by exploiting the diverging outputs of distinct \nn{} models. We
%% formulate this attack method as an optimization problem to which we
%% apply the \emph{Hill Climbing}~\ourcite{Russell2009}.

%\subsection{Workflow}

%---------------------------------------------
\subsection{Attack formulation}
%---------------------------------------------

Technically, \tname{} systematically searches the input space to (i) maximize the
output/output difference while (ii) minimizing the input/input
difference. The output/output difference consists of the difference
between the outputs produced by each \nn{}. For example, for
classification problems, \nn{}s often reports probability
distributions indicating the likelihood of the input belonging to a
given class. In those cases, \tname{} maximizes the difference between
the two distributions reported. Simultaneously, \tname{} minimizes the
difference between the perturbed input and the original input. 
%More precisely,
%\tname{} minimizes the Euclidean distance (\aka{} L2 norm) between the
%input and output.

%% \tname{} uses
%% the input $x$ as seed to search for adversarial inputs. In \stp{1},
%% \tname{} makes small perturbations in $x$, referred to {x'}.  In
%% \stp{3} \tname{} queries the two \nn{}s. If they produce different
%% classifications for the input, \tname{} terminates reporting \tm{x'}
%% as an adversarial input. However, if the \nn{} agree in the
%% classification, \tname{} returns to \stp{1} and continuously repeating
%% \textbf{Steps 1-3} until the \nn{} produce divergent classifications,
%% finally terminating.
%

%Let us consider that $f_1$ and $f_2$ denote functions associated with
%two \nn{}s models, $M_1$ and $M_2$, respectively. Given an input $x$,
%the goal of the attack is to find candidate inputs $x'$ that maximize
%the difference in the outputs of $f_1$ and $f_2$ while keeping the
%difference between $x$ and $x^\prime$ below a specified
%threshold. Formally, 
Formally, given two functions $f_1$ and $f_2$ and an input $x$, the optimization problem can be described as:
\begin{equation}
 \centering
 \label{equ:diffmax}
 \begin{array}{rl}
 \max_{x' \in {\Bbb R} ^d} \|f_1(x') - f_2(x')\|_1 \text{ s.t.} &  \|x - x'\|_p \leq \epsilon \text{ , } \arg{}\max{} f_1(x) \neq  \arg{}\max{} f_2(x').
 \end{array}
\end{equation}
Intuitively, it is to find an adversarial example $x'$ that maximizes the difference between outputs from $f_1$ and $f_2$, such that $x'$ is not too far away from $x$, and there exists a misclassification.

Consider the example illustrated in Figure~\ref{fig:example} and the formalization in Eq.~\eqref{equ:diffmax}. One could apply an optimization algorithm to randomly add noise patterns to the street road. This means that noises (or combinations of noises), which maximize the difference between the output of \nn{1} and the output of a \nn{2} for the perturbed street road, will be kept until the search algorithms find a feasible solution.

\subsection{Hill Climbing (HC) optimization}
%---------------------------------------------

In order to solve the optimization problem in Eq.~\eqref{equ:diffmax},  Hill Climbing (HC)~\ourcite{Russell2009}, a mathematical optimization technique suited for local search problems, is chosen. HC is an iterative algorithm that starts with an arbitrary solution for a given problem, and at each iteration, it adds incremental alterations to the solution. The core components of HC are described below:
\begin{itemize}
\item \noindent\textbf{Selection.} A new solution is generated by adding random mutation to a previously generated solution.
\item \noindent\textbf{Evaluation.} The evaluation function calculates a score that measures the quality of a solution candidate.
\item \noindent\textbf{Termination.} The algorithm terminates if the solution has converged, i.e., the algorithm is unable to produce new solutions that are different from the previous solution.
\end{itemize}

% ------------------------------------
% Hill Climbing
%-------------------------------------
\begin{algorithm}[!h]
\caption{\label{alg:hillClimb} Hill Climbing algorithm for difference inducing adversarial examples generation.}
\DontPrintSemicolon

\KwData{Input original example $x$; two target models \nn{1} and
\nn{2} as black-box functions $f_{1}$ and $f_{2}$, respectively;
maximum number of allowed iterations $T$; a constant $c$ to re-scale the \lnorm{} value}
 \KwResult {Output an adversarial example $x'$ if any, or the
best solution it could find before termination.}\;

\SetKwProg{Fn}{Function}{:}{}
\Fn{Selection($x'$)}{%
    \nl pixelIndex~$\gets$~rand(0,~len(x'))\;
    \nl $x'[pixelIndex]~\gets~rand(0, 255)$\;
    % \nl $x'[pixelIndex]~\gets~\Phi()$\;
    \KwRet x\;
}

\Fn{Evaluation($x'$)}{%
    \KwRet $\Omega(x, x')$\;
    % \KwRet $abs(f_1(x') - f_2(x')) - norm(x'- x)/norm(x')$\;
}

\nl sol $\gets$ Selection(x)\;
\nl score $\gets$ Evaluation(x)\;
\nl i $\gets$ 0\;
\While{$\neg Termination(sol, i)$}{
    \nl newSol $\gets$ Mutation(sol)\;
    \nl newScore $\gets$ Evaluation(newSol)\;
    \If{newScore > score}{
        \nl sol $\gets$ newSol\;
        \nl score $\gets$ newScore\;
    }
    \nl i $\gets$ i+1\;
 }
 \nl x' $\gets$ sol\;
\end{algorithm}

In Algorithm~\ref{alg:hillClimb}, we present our implementation of the HC algorithm for \tname{} attack. This algorithm receives as input four parameters as follows. a) $x$ is the original image. b) $y$ is the true label of  $x$. c) $f_{1}$ and $f_{1}$ are  two NN models \nn{1} and \nn{2}. d) $T$ is the maximum number of iterations the algorithm is allowed to have. e) $c$ is a constant factor that we can use to re-scale the \lnorm{} value, and it is used during the evaluation step (see, Eq.~\eqref{eq:fitness}). This is important because we need to make sure the \lnorm{} and the score derived from the difference between the \nn{} are on the same scale to avoid that algorithm gives to much importance to the perceptual aspect of our optimization (i.e., minimizing the difference between the images), ignoring the differential aspect. On the other hand, if we set $c=0$ we can make the algorithm ignore the perceptual aspect of the optimization and focus only on maximizing the differential behavior of the \nn{}s. The basic intuition behind this is that the algorithm can only succeed if it is able to find difference inducing solutions, which is achieved by maximizing divergent behavior on the \nn{}, while the \lnorm{} forces the algorithm to find solutions that are closer to the original inputs. If that is not a need we can flex the importance of this aspect of the attacks. The mutation function is responsible for adding pixel-level noises/perturbations to a given image x; it randomly selects a single pixel from the image and modifies its original value by a new randomly generated value.  A generated solution is then evaluated by calculating a score using Eq.~\eqref{eq:fitness}.

\begin{equation}
    \label{eq:fitness}
    \begin{array}{l}
    %\Omega (x^\prime, y) = \|f_0(x^\prime, y) - f_1(x^\prime, y)\|_1 = \|P_0(y|x^\prime) - P_1(y|x^\prime)\|_1
    \displaystyle \Omega (x, x') = abs(f_1(x') - f_2(x')) - c * norm(x'- x)
    \end{array}
\end{equation}

%---------------------------------------------------------------------------------
\section{Evaluation}
\label{sec:experimentation}
\label{sec:eval}
%---------------------------------------------------------------------------------

We organized the evaluation based on whether or not a technique
leverages code of the \nn{} to obtain adversarial inputs (\ie{},
white- versus black-box) and whether or not a technique takes multiple
\nn\ models on input (\ie{}, differential versus
non-differential). Intuitively, \emph{our goal} is to understand the
impact of each of these aspects on the performance of an adversarial
input generation technique.

We pose the following questions:

\newcommand{\rqone}{How \tname{} compares with White-box Differential
  techniques? (Section~\ref{eval:whitebox-DAEGEN})}

\newcommand{\rqtwo}{How \tname{} compares with Black-box
  Non-differential techniques? (Section~\ref{eval:blackbox-DAEGEN})}

\begin{itemize}
\item{\textbf{RQ1.}} \rqone
\item{\textbf{RQ2.}} \rqtwo
\end{itemize}

We conducted all experiments on a machine with two Intel(R) Core(TM)
i7-8750H CPU @ 2.20GHz processor, 2208 Mhz, 6 Core(s), 12 Logical
Processor(s), 32 GB of DDR4 memory, NVIDIA GeForce GTX 1070 GPU. We
used Python 3.5.6, PyTorch 1.0.1, Torchvision 0.2.23, Keras 2.0.8,
Tensorflow 1.10.0 and Windows 10 Pro OS.

%We do not set limits on time or memory.

The tool implementing \tname{} and the scripts to reproduce our experimental results are publicly available at the following link \textbf{\url{https://github.com/DAEGEN/DAEGENAttack}}.

\subsection{Metrics}
\label{sec:metrics}
%%%%%%%%%%%%%%%%%%%%%%%%%%%%%%%%%%%%%

We evaluated techniques in three dimensions:

\begin{enumerate}

\item \emph{effectiveness}
  determines how successful a technique is in generating adversarial
  examples;
  
\item \emph{precision} determines the ability of generated examples to
  fool machines and humans;
  
\item \emph{efficiency} determines how computationally efficient a
  technique is.
\end{enumerate}  

The following sections explain the metrics used to assess
\textit{effectiveness}, \textit{precision}, and \textit{efficiency}.

%% ---conceptually, the smaller the difference
%% between adversarial and non-adversarial inputs, the higher the chances
%% the attack will be useful. Lastly, 3) 
%% We evaluated techniques on three dimensions: \emph{effectiveness},
%% \emph{precision}, and \emph{efficiency}. Effectiveness refers to the
%% ability of generating adversarial inputs, precision refers to the
%% quality of these inputs, and efficiency refers to the computational
%% cost for generating those inputs.

%%%%%%%%%%%%%%%%%%%%%%%%%%%%%%%%%%%%%
%% In literature, three dimensions are often mentioned regarding how to evaluate black-box adversarial attacks: 

%% \begin{itemize}
%% \item \textbf{Attack Success Rate (ASR)} indicates how often the optimization algorithm finds an adversarial example.
%% \item \textbf{Number of queries} indicates how many queries were required by the optimization algorithm to find an adversarial example.
%% \item \textbf{Perturbations norms}, for example, 
%% \lnorm{} norm 
%% \end{itemize}

%%%%%%%%%%%%%%%%%%%%%%%%%%%%%%%%%%%%%
\subsubsection{Effectiveness}
\label{sec:eval-effectiveness}
%%%%%%%%%%%%%%%%%%%%%%%%%%%%%%%%%%%%%

%\vspace{1ex}\noindent

We used the Differential Success Rate \textbf{(DSR)} as proxy for \emph{effectiveness}. This metric measures the relative number of DIAEs (see Section~\ref{sec:introduction}), which a given technique produces on a pair of models. In the following, we define how we computed DSR for differential and non-differential techniques. These definitions differ because non-differential techniques take one \nn{} model on input, in contrast with differential techniques, which take two. \textbf{DSR of Differential Techniques.}  Let $M$ define the kind of a network model, and let $X$ define the set of inputs (\eg{}, the set of images). We define a differential technique as a function $t$:~$M\times M\times X \rightarrow \mathbb{Z}_2 = \{0,1\}$, where $1$ indicates that the technique was successful in producing an adversarial example, while $0$ indicates otherwise. That function takes two classification models in $M$ and seed in $X$ as input and produces an output indicating success or failure. In successful attacks, for any pairs of models $m_1, m_2 \in M$, the following proposition must hold $m_1(x)\neq{}m_{2}(x)$, for any seed input $x \in X$. The Differential Success Rate (DSR) of a \emph{differential   technique} $t$ is obtained by comparing the outputs of any two models $m_1, m_2 \in M$ on a given set of seed inputs $X_0 \subseteq X$. More formally, we define the DSR of a technique $t$ with respect to two models $m_1$ and $m_2$ and inputs $X_0$ as $DSR(t, m_1, m_2, X_0) = \frac{1}{\vert X_0\vert}\times\sum_{x}^{X_0}{t(m_1, m_2, x)}$.  $X_{0}$, referred to as \textbf{DSR}(t, $X_{0}$), is obtained by computing the average $DSR$ across all pairs of models in $M$. \textbf{DSR of Non-Differential Techniques.}~ We included non-differential techniques in the evaluation. Unfortunately, the   DSR formulation above is inadequate for these techniques, as they take only one model on input. To circumvent that issue, we defined a   differential technique $p$, based on its non-differential technique   $t$, to decide if the adversarial input produced by $t$ is a DIAE   (see Section~\ref{sec:introduction}). Intuitively, $p$ checks, for a   given pair of models $\langle{}a, b \rangle$, whether or not the adversarial input produced by a (non-differential) technique for the model $a$ results in a different output when fed to $b$. The computation of DSR is as described for a differential technique.

%% More
%% formally, a non-differential technique $t~:~M~\times~X~\times~Y
%% \rightarrow <\mathbb{Z}_2, x'>$ takes one classification model in $M$,
%% a seed in $X$, and the respective label of the seed in $Y$ as inputs
%% and produces a tuple containing a binary value indicating whether the
%% attack was successful or not and an adversary image $x'$\Mar{shouldn't
%%   be the type/domain?}, also indicating success or failure. Assuming
%% $y_x \in Y$ is the label of the seed $x \in X$, for any model $m \in
%% M$, the following proposition must hold $m(x) \neq y_x$ for a
%% successful attack. 

%% For any pair of models $\langle{}m_1, m_2 \rangle$, $p$ takes
%% the the first model in the pair and a seed image, invokes $t$ on that
%% input

%% $<z, x'>$ where $z \in \{0, 1\}$, $x'$ is a difference inducing input
%% if $z=1$ and $m_1(x') != m_2(x')$.

%% Formally,
%% $p~:~t~\times~M~\times~M~\times~X~\times~Y \rightarrow \mathbb{Z}_2$
%% takes two classification models from $M$, a non-differential technique
%% $t$, a seed in $x$, and the respective label of the seed in $Y$ as
%% inputs, and produces a binary output where 1 indicates the technique
%% $t$ successfully generated a difference inducing input, and 0
%% indicates otherwise. 

%%%%%%%%%%%%%%%%%%
\subsubsection{Precision} 
%%%%%%%%%%%%%%%%%%
Given the impossibility to automatically quantify the ability of humans to perceive changes in a pair of images~\ourcite{Hafemann2018}, it is common practice to adopt metrics of image similarity as a proxy of perception. We used the Euclidean distance between two images -- the \lnorm{} norm -- to quantify image similarity. This norm has been consistently used in prior related work.

%%%%%%%%%%%%%%%%%%%
\subsubsection{Efficiency} 
%%%%%%%%%%%%%%%%%%%
We used time and number of queries
submitted to the models to measure efficiency, which has been shown to
be important in some domains~\ourcite{Bhambri2019ASO}.

%% \textbf{The number of queries} is a proxy of efficiency. It indicates how many times the attack method had to send an input to the model under attack and to note the observations made before a successful adversarial example was produced. Minimizing the number of queries reduces the costs (\eg{}, time, money, resources) needed to craft an adversarial example, being, therefore, a vital aspect of an efficient adversary technique.

%%%%%%%%%%%%%%%%%%%%%%%%%%%%%%%
\subsection{White-box (Differential) Techniques}
\label{eval:whitebox-DAEGEN}
%%%%%%%%%%%%%%%%%%%%%%%%%%%%%%%

This section elaborates on the comparison of \tname{} with white-box techniques.

%%%%%%%%%%%%%%%%%%%%%%%%%%%%%%%
\subsubsection{Comparison Techniques}
\label{sec:comparison-techs-wb}
%%%%%%%%%%%%%%%%%%%%%%%%%%%%%%%
We selected two recently-proposed popular white-box differential techniques for comparison -- DeepXplore~\ourcite{pei-etal-sosp2017} and DLFuzz~\ourcite{guo-etal-fse2018}. DeepXplore introduced the concept of neuron coverage criterion and was the first technique to use differential testing in this domain. It uses multiple (similar) \nn{}s to find discrepancies in results. The approach exploits the differential behavior of a set of \nn{}s to generate adversarial input images. The adversarial input generation problem is formalized as a joint optimization problem, where the following objectives are considered: 1)~to maximize output divergence and 2)~to maximize neuron coverage. A gradient ascent search strategy is used to create perturbations on inputs to maximize both objectives. We considered three configurations of DeepXplore for comparison: Light, Occl, and Blackout. Similar to \dx{}, \df{} is also differential. However, \df\ does \emph{not} require additional model(s) to find adversarial attacks. Instead of looking for inconsistent behavior across many \nn{} models, it looks for inconsistent classification within a single \nn{} model. The attack method assumes that the first classification given by the classifier for a given seed input is correct and takes it as ground truth. An adversarial attack is reported if the technique can produce similar inputs with different classification.

%%%%%%%%%%%%%%%%%%%%%%%%%%%%%%%
\subsubsection{Datasets and Models} 
%%%%%%%%%%%%%%%%%%%%%%%%%%%%%%%
We used two image datasets in the evaluation: MNIST~\ourcite{Lecun1998} and Driving~\ourcite{udacity-challenge}. MNIST is a dataset of handwritten digits, which contains $70$K grayscale $28$$\times$$28$ (in pixels) images, divided into ten classes, \ie{}, digits $0$--$9$. Driving is a dataset made available by the Udacity self-driving car challenge. It contains $5,614$ images captured by a self-driving car labeled with the steering wheel angle that should be applied by the driver at the time each image was captured.

%%  and later used by
%% Alzantot \etal{} in the evaluation of
%% GenAttack~\ourcite{Alzantot2019}

%\begin{wraptable}[13]{r}{0.22\textwidth}
\begin{table}[!h]
  \small
  \caption{\label{tab:modelxmethodxWBtechniques}Datasets and Models
    used for comparison of \tname{} with White-Box techniques.}
  \vspace{-1ex}
  \begin{tabular}{ccc}
    \toprule
    \multicolumn{1}{c}{Datasets} & Model (Source) & Used at \\
    \midrule
    \multirow{3}{*}{MNIST} & LeNet1~\ourcite{Lecun1998} & \multirow{3}{*}{\citet{pei-etal-sosp2017,guo-etal-fse2018}} \\ 
    &  LeNet4~\ourcite{Lecun1998} &  \\ 
    & LeNet5~\ourcite{Lecun1998} &  \\
    \midrule
    %% \multirow{7}{*}{??} & \textbf{ResNet34}~\ourcite{He2015DeepRL} & \ourcite{Huang2020} \\ 
    % \multirow{3}{*}{ImageNet} & ResNet50~\ourcite{He2015DeepRL} & \multirow{3}{*}{\citet{pei-etal-sosp2017,guo-etal-fse2018}} \\
    % & VGG16~\ourcite{Simonyan2014deep} & \\ 
    % & VGG19~\ourcite{Simonyan2014deep} &  \\ \midrule
    % Dave-orig, %% which one actually proposed the model?
    \multirow{3}{*}{Driving} & Dave-orig~\ourcite{Bojarski-etal-JM16} & \multirow{3}{*}{\citet{pei-etal-sosp2017}} \\
    & Dave-norminit~\ourcite{Simonyan2014deep} &  \\ 
    & Dave-dropout~\ourcite{Simonyan2014deep} & \\ 
    %% & \textbf{VGG19bn}~\ourcite{Simonyan2014deep} & \ourcite{Huang2020}  \\ 
    %% & \textbf{DenseNet121}~\ourcite{Huang2016DenselyCC} & \ourcite{Huang2020}  \\ 
    %% & \textbf{MobileNetV2}~\ourcite{Sandler2018MobileNetV2IR} &
    %% \ourcite{Huang2020} \\
    \bottomrule
  \end{tabular}
\end{table}

Table~\ref{tab:modelxmethodxWBtechniques} shows the datasets and
models we used. We focused on datasets and models 1) used in the
evaluation of both \dx\ and \df\ and 2) used in the autonomous-vehicle
domain.

%These models and images were considered in
%the evaluation of \dx{} but not in the evaluation of \df{}.

%%%%%%%%%%%%%%%%%%%
\subsubsection{Setup}
%%%%%%%%%%%%%%%%%%%

%% to produce an attack for a
%% given input. This value is higher than the value used by other
%% black-box techniques. As we explain further, we do not limit white-box
%% techniques based on any constraints. As a result, we decided to give
%% \tname{} a comfortable query budget. Additionally, the query number is
%% not relevant in the white-box comparison, but could impact \tname{}
%% overall performance.

We set a budget of $50$K queries on \tname{} to give it enough time to generate adversarial examples. Recall that \tname{} stops when the first adversarial example is produced. In that case, the budget is not reached. Such a high number of queries enables us to understand the actual cost of \tname{} better. \dx{} and \df{}, unlike \tname{}, can produce multiple adversarial examples for a given input. To fairly compare these techniques with \tname{}, we recorded only the first adversarial example they generate for a given input seed. Note that number of attacks per input is irrelevant for DSR, which focuses on whether or not an attack was successful.

We followed usage instructions from \dx{}~\cite{dxrepository} and \df~\cite{dlfrepository}; we did not change any part of their code. \dx{} and \df{} offer command-line parameters to stop when the first example is found. This methodology ensures that when any of these tools stop, they either succeed in generating adversarial inputs or fail due to their limitations.

%---------------------------------
\subsubsection{{Results}}
\label{subsec:results}
%---------------------------------

Table~\ref{tab:dsr} summarizes the results of each technique. We
grouped results by datasets
(see~Table~\ref{tab:modelxmethodxWBtechniques}). Recall that
\df\ cannot handle the Driving dataset as is and we chose not to
change the code~(see~Section~\ref{sec:comparison-techs-wb}). Observe
that, for ``DSR'' (effectiveness), the higher the value the better,
whereas for ``Avg. $L_2$'' (precision) and ``Avg. Time (s)''
(efficiency), the lower the value the better.

\begin{table}[h!]
  \small
  \caption{\label{tab:dsr}Average results comparing \tname{} and White-Box
    techniques: \dx\ (three variants)~\ourcite{pei-etal-sosp2017} and
    \df~\ourcite{guo-etal-fse2018}.}
\setlength{\tabcolsep}{12pt}  
\begin{tabular}{c l c c c}
  \toprule
  Dataset & Technique & DSR & \lnorm{} & Time (s)\\
  \midrule
  \multirow{5}{*}{MNIST} & DeepXplore light & 0.732 & 14.034 & 5.573\\ 
  & DeepXplore occl & 0.831 & 5.029 & 6.477 \\ 
  & DeepXplore blackout & 0.624 & \cellcolor{lightgray}1.041 & 7.321\\
  & DLFuzz & 0.918 & 2.611 & 5.205   \\
  & \tname{} & \cellcolor{lightgray}1.0 & 6.691 & \cellcolor{lightgray} 0.4 \\
  \midrule
  \multirow{4}{*}{Driving} & DeepXplore light & 0.366 & 90.055 & 11.006 \\ 
  & DeepXplore occl & 0.927 & 60.680 &8.683\\ 
  & DeepXplore blackout & 0.070 & 25.765 & 9.271\\
  & \tname{} & \cellcolor{lightgray}1.0 & \cellcolor{lightgray}7.082 & \cellcolor{lightgray}4.939   \\
  \bottomrule
\end{tabular}
\end{table}

% \begin{table}[h!]
%   \small
%   \caption{\label{tab:dsr}Average results comparing \tname{} and White-Box
%     techniques: \dx\ (three variants)~\ourcite{pei-etal-sosp2017} and
%     \df~\ourcite{guo-etal-fse2018}.}
% \setlength{\tabcolsep}{12pt}  
% \begin{tabular}{llrrr}
%   \toprule

%   \bottomrule
% \end{tabular}
% \end{table}

% \begin{table}[h!]
%   \small
%   \caption{\label{tab:dsr}Average results comparing \tname{} and White-Box
%     techniques: \dx\ (three variants)~\ourcite{pei-etal-sosp2017} and
%     \df~\ourcite{guo-etal-fse2018}.}
% \setlength{\tabcolsep}{12pt}  
% \begin{tabular}{lrrr}
%   \toprule

%   \bottomrule
% \end{tabular}
% \end{table}

Results indicate that the \dx\ blackout combination was the one that performed best in the MNIST, where it produced examples with lower \lnorm. However, that combination failed to produce examples in several cases. Overall, \df{} performed well on the MNIST dataset since it produced adversarial examples in 91.8\% of the cases, and the examples have relatively low \lnorm\ values.
Results show that \tname{} generated adversarial examples in every case and that it was more efficient than every other technique. The average \lnorm\ norm of (the examples produced by) \tname{} is not always superior to other techniques' \lnorm. For example, three techniques on the MNIST dataset produce examples with lower average \lnorm\ than \tname. The reason is that the setup we used for \tname{} weighs efficiency (\ie{}, execution time) higher than precision (\ie{}, \lnorm). In principle, all techniques could produce examples with lower \lnorm\ by executing each query until reaching execution bounds instead of stopping on the first adversarial example produced. However, that setup would lead to higher execution times, as we did. We elaborate and discuss the importance of \lnorm\ in the following.

\vspace{0.5ex} \textbf{Effectiveness.}~Table~\ref{tab:dsr-breakdown} shows the breakdown of DSR per \nn\ model. The columns under ``DSR'' show the DSR of a technique for a given pair of the dataset (column ``Dataset'') and model.  Note that \dx{} and \df{} perform relatively well when considering DSR alone. The highlighted cells show the best DSR obtained below 1.0. Overall, \df{} performed well on the MNIST dataset, whereas \dx{} occl performed well on Driving. Recall that \tname{} can produce adversarial examples in every case. 

%% Compared with \df, \tname{} produced images with higher L2 values, but
%% it obtained results slightly faster and was successful in 5\% more
%% cases.

%% Compared with the configurations of \dx,
%% \tname{} was superior in almost every metric, except for L2 for
%% \dx\ blackout, which produced adversarial attacks with the smallest
%% average L2. Note, however, that \dx\ blackout was only successful in
%% 38\% of the cases on average (as per DSR).

\begin{table}[h!]
  \small
  \caption{\label{tab:dsr-breakdown}DSR breakdown.}
\setlength{\tabcolsep}{12pt}  
\begin{tabular}{cccccc}
  \toprule
 Dataset & Technique & \multicolumn{3}{c}{DSR}\\
 \midrule
  &  & LeNet1 & LeNet4 & LeNet5 \\
  \midrule
  \multirow{5}{*}{MNIST} & DeepXplore light & 0.697 & 0.703 & 0.796 \\ 
  & DeepXplore occl & 0.839 & 0.845 & 0.806 \\ 
  & DeepXplore blackout & 0.623 & 0.596 & 0.650\\
  & DLFuzz & 0.933 & 0.893 & 0.929   \\
  & \tname{} & \cellcolor{lightgray}1.0 & \cellcolor{lightgray}1.0 & \cellcolor{lightgray}1.0   \\
 
 \midrule
  &  & Dave-Orig & Dave-Drop & Dave-Norm \\
 \midrule
  \multirow{4}{*}{Driving} & DeepXplore light & 0.212 & 0.451 & 0.435 \\ 
  & DeepXplore occl & \cellcolor{lightgray}1.00 & \cellcolor{lightgray}1.00 & 0.782\\ 
  & DeepXplore blackout & 0.005 & 0.091 & 0.115\\
  & \tname{} &  \cellcolor{lightgray}1.0 &  \cellcolor{lightgray}1.0 &  \cellcolor{lightgray}1.0   \\
  \bottomrule
\end{tabular}
\end{table}

\begin{figure*}[h!]
  \centering
  \begin{subfigure}{\textwidth}
    \centering
    \includegraphics[width=0.24\linewidth]{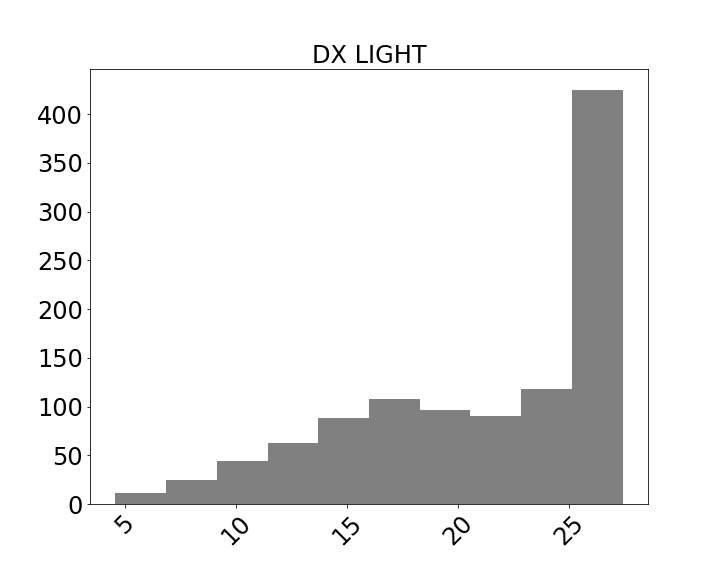}
    \includegraphics[width=0.24\linewidth]{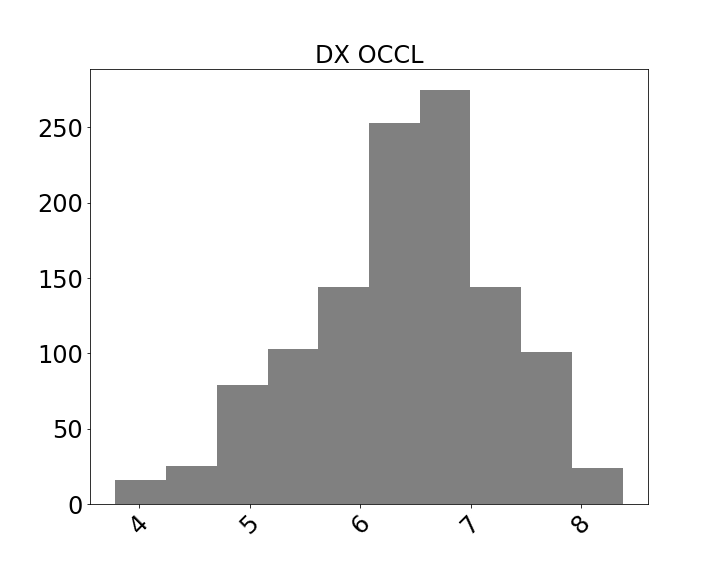}
    \includegraphics[width=0.24\linewidth]{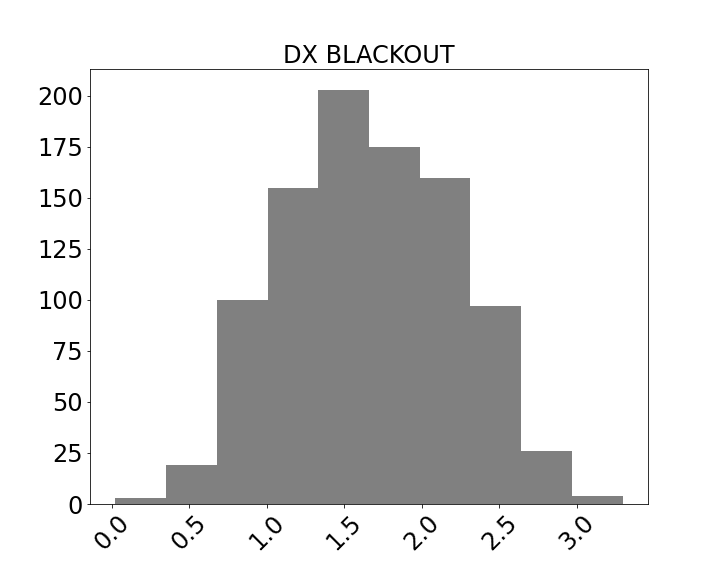}
    \includegraphics[width=0.24\linewidth]{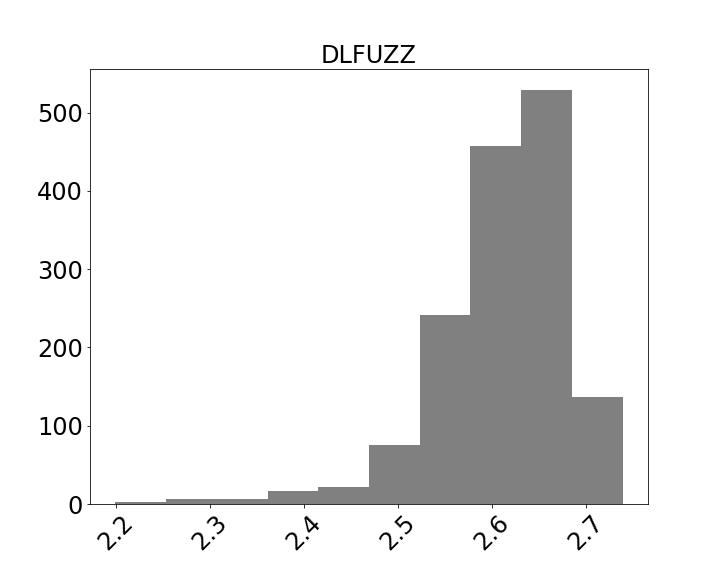}
    \includegraphics[width=0.24\linewidth]{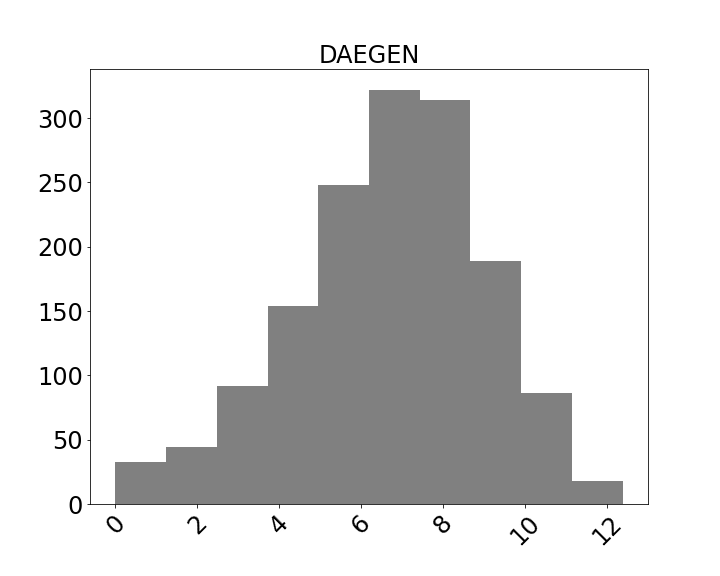}
    \caption{MNIST}
  \end{subfigure}
  \begin{subfigure}{\textwidth}
    \centering
    \includegraphics[width=0.24\linewidth]{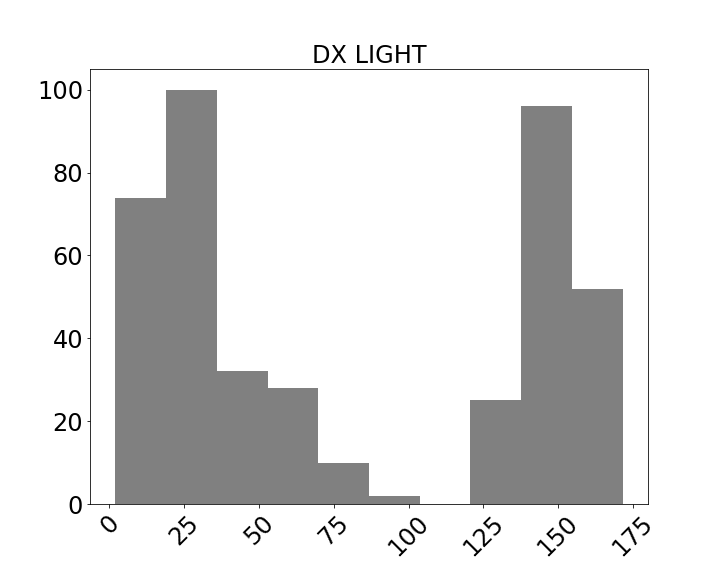}
    \includegraphics[width=0.24\linewidth]{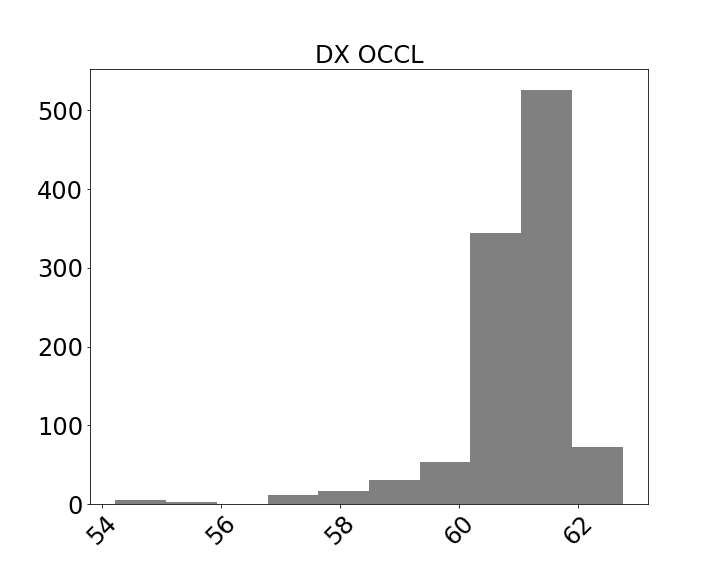}
    \includegraphics[width=0.24\linewidth]{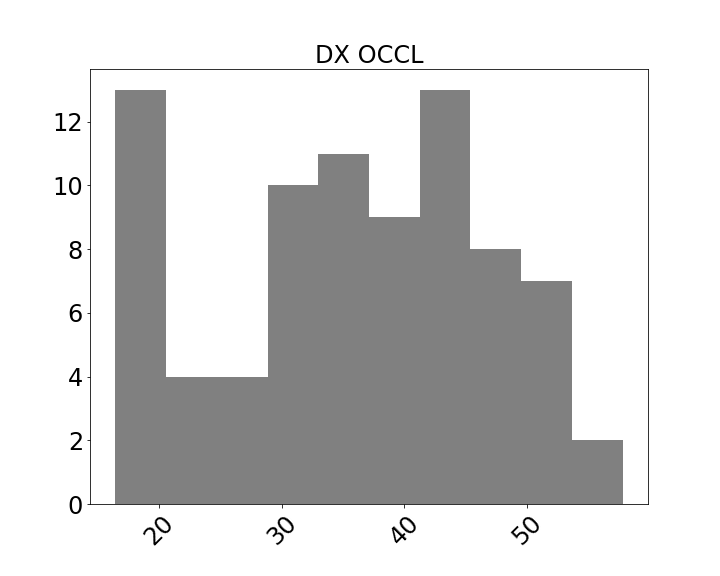}
    \includegraphics[width=0.24\linewidth]{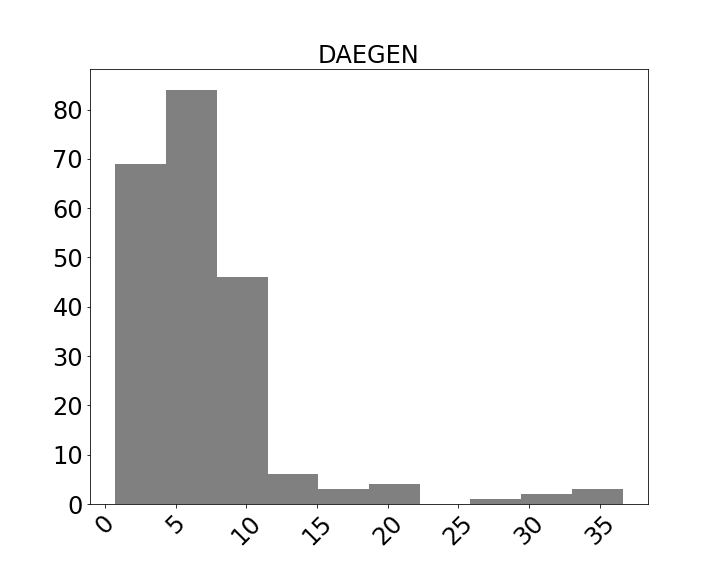}
    \caption{Driving}
  \end{subfigure}
    \caption{\lnorm\ distributions per dataset.}
    \label{fig:l2distributionsWB}
\end{figure*}

\vspace{0.5ex} \textbf{Precision.}~
Figure~\ref{fig:l2distributionsWB} shows the distributions of
\lnorm\ associated with each technique and dataset.  Each bar
indicates the number of adversarial examples generated (y axis) for a
given range of \lnorm\ values (x axis).
Overall, the results of \dx\ on \lnorm\ are not consistent. For example, for the MNIST dataset, \dx{} occl and blackout produced examples with lower \lnorm\ values than \tname. On the Driving dataset, however, every configuration of \dx\ produced examples with high \lnorm\ values. Considering the distribution of \df\ on the MNIST dataset, note that most examples have \lnorm\ in the 2.6-2.7 range, which is relatively low.

\tname{} produced examples in both datasets with similar values for \lnorm.
Compared to the \dx\ combinations on the Driving dataset, \tname{}
produced examples with considerably smaller \lnorm\ values. However,
it produced examples with high \lnorm\ values on the MNIST
dataset. Three reasons justify that result. First, the number of
queries \tname{} uses to produce an attack influences on \lnorm{}---a
higher number of queries produces more noise on an image. We observed
that the Pearson's correlation between the number of queries and
\lnorm{} was, in fact, moderate ($\sim$0.6). Second, \tname{}
generates noise with values ranging in the 0-255 interval. If a noised
pixel has a value that is distant from the original pixel value (e.g.,
the original pixel value is 0 and the noised pixel value is 255), this
difference is likely to result in a higher \lnorm{} value. In
contrast, \dx{} and \df{} use other criteria to determine the range of
perturbation. For example, \dx{} uses a loss function based on the
gradients of the \nn\ to determine the range of perturbation. Third,
\tname{} stops when the search finds the first adversarial example for
a given seed input.

%% \Mar{what is a loss
%%   function?} \Joao{it is the cost associated with their objective
%%   function.  The objective function is differentiable, so they
%%   calculate the gradients and take steps proportional to the positive
%%   of the gradients (gradient ascent optimization)}

\begin{figure}[h!]
    \centering
    \includegraphics[width=0.22\linewidth]{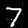}
    \includegraphics[width=0.22\linewidth]{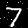}
    \includegraphics[width=0.22\linewidth]{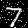}
    \includegraphics[width=0.22\linewidth]{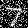}
    \includegraphics[width=0.22\linewidth]{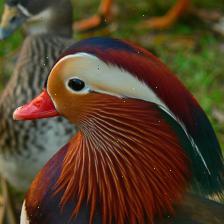}
    \includegraphics[width=0.22\linewidth]{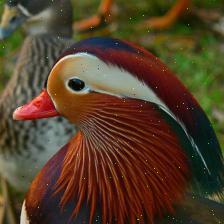}
    \includegraphics[width=0.22\linewidth]{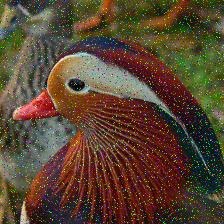}
    \includegraphics[width=0.22\linewidth]{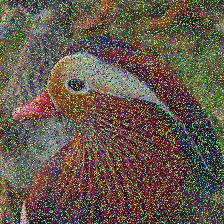}
    \includegraphics[width=0.22\linewidth]{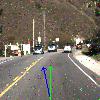}
    \includegraphics[width=0.22\linewidth]{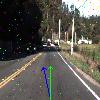}
    \includegraphics[width=0.22\linewidth]{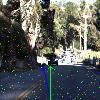}
    \includegraphics[width=0.22\linewidth]{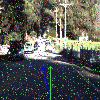}
    \caption{Adversarial examples created for a given seed image with increasing L2 values.}
    \label{fig:increasingl2}
\end{figure}

It is important to note that it remains an open research problem the
identification of a threshold for a given norm (\eg{}, $L_2$) below
which a human would unlikely be able to detect differences in a pair
of images~\ourcite{Hafemann2018}. In other words, high $L_2$ values do
not necessarily invalidate an adversarial
input. Figure~\ref{fig:increasingl2} shows inputs that \tname{}
creates, with increasing values of \lnorm{} from left to right, to
illustrate that. 

The first row shows example images from the MNIST dataset with \lnorm\ values of 1, 2, 5 and 10, respectively. The impact of perturbations as the norm increases is noticeable in this case. In contrast, the impact of the perturbations is less noticeable on the \inet{} example~\ourcite{Deng2009}, on the second row, and the Driving example, on the third row. For comparison, the four images of a drake duck have associated \lnorm\ values of 5, 10, 50, and 100, respectively. To sum up, human perception is not a function of $L_2$ alone. The resolution of the image also plays a role in that.

%% \Mar{agree? if yes, can you add
%%   a sentence about resolution of these images?} \Joao{If by resolution
%%   you mean the images dimensions (\eg{}, 3x224x224, 3x100x100, 28x28),
%%   we can say that, the less pixels a image have, the more it will be
%%   sensible to noises. That is why the l2 norms for \inet{} is in a
%%   different scale.} 

%% \Mar{Can we say results, although not
%%   optimal, are ``good enough''? does it make sense to add a sentence
%%   saying that we inspected a sample of these images and could not
%%   identify differences?}

\begin{figure*}[h]
  \centering
  \begin{subfigure}{\textwidth}
    \centering
    \includegraphics[width=0.24\linewidth]{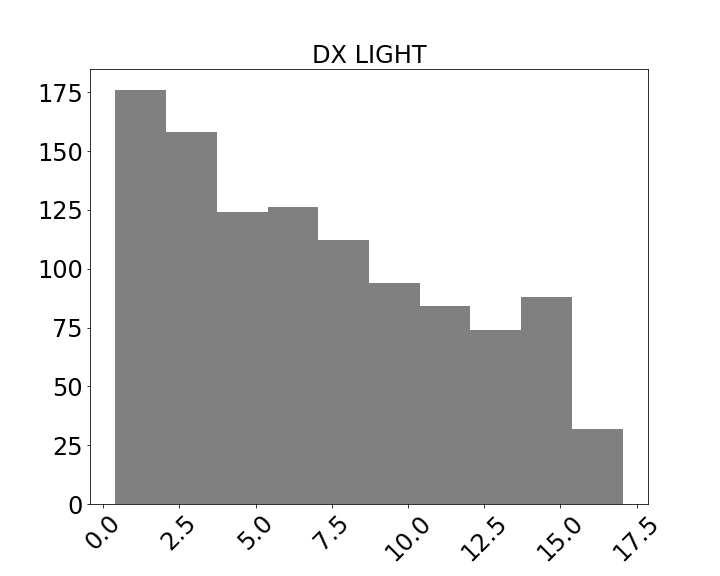}
    \includegraphics[width=0.24\linewidth]{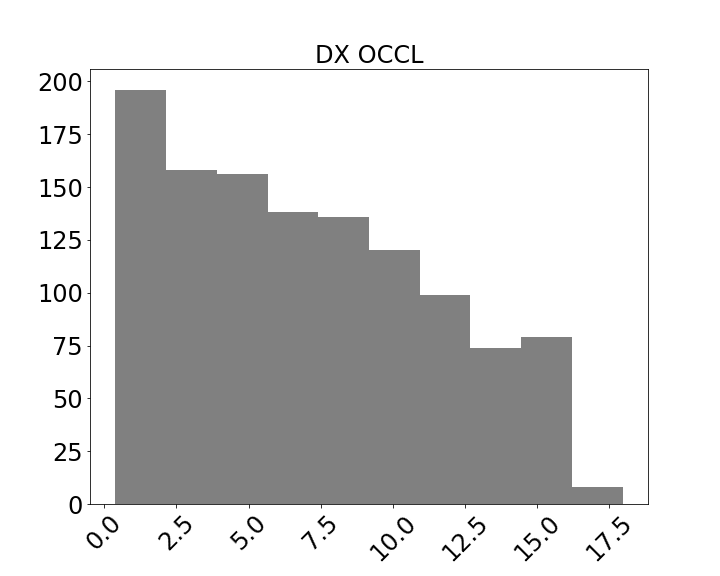}
    \includegraphics[width=0.24\linewidth]{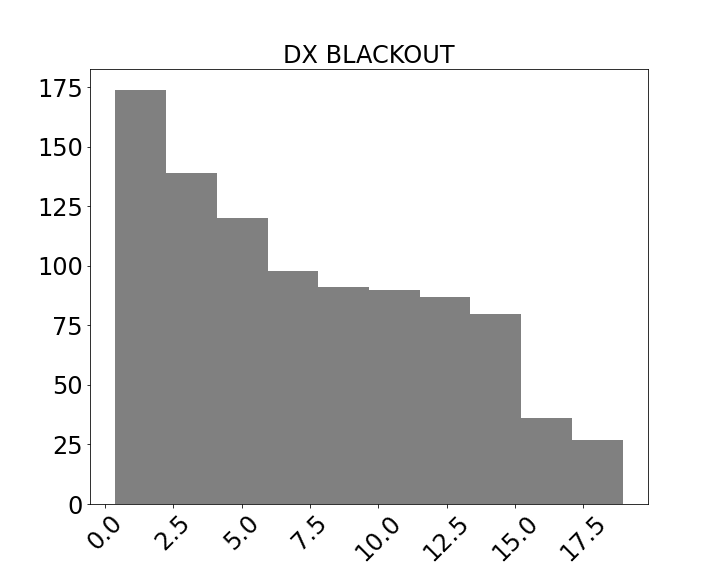}
    \includegraphics[width=0.24\linewidth]{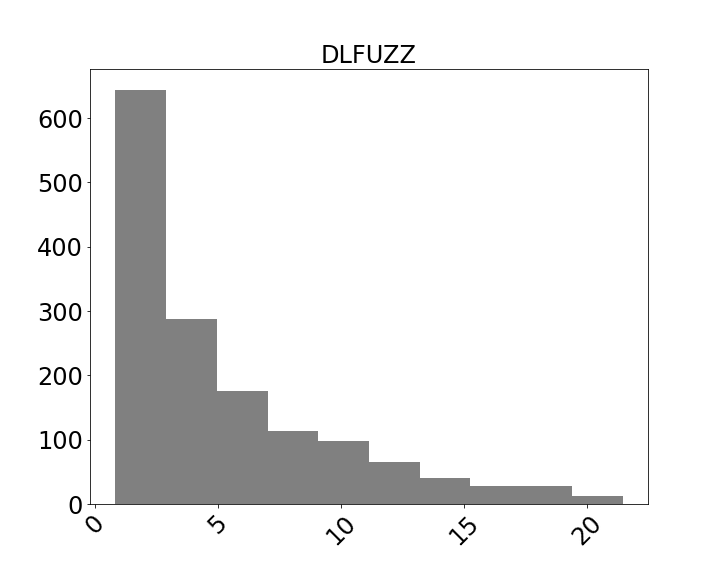}
    \includegraphics[width=0.24\linewidth]{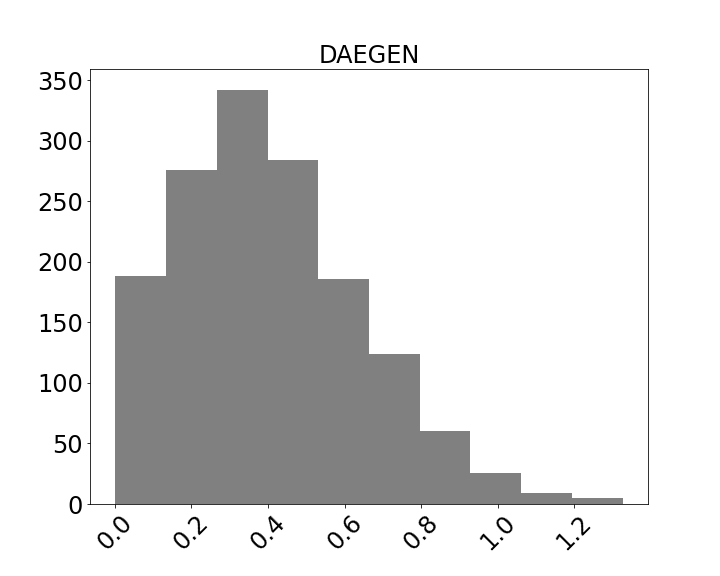}
    \caption{MNIST}
  \end{subfigure}
  \quad\quad
  \begin{subfigure}{\textwidth}
    \centering
    \includegraphics[width=0.24\linewidth]{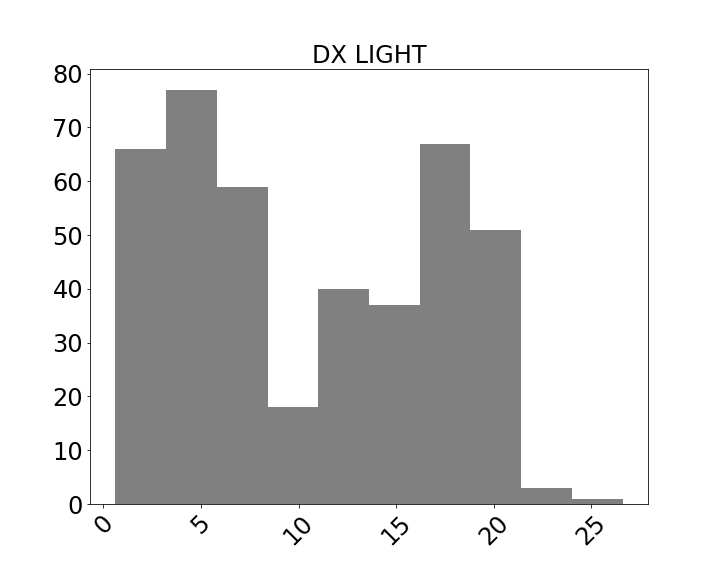}
    \includegraphics[width=0.24\linewidth]{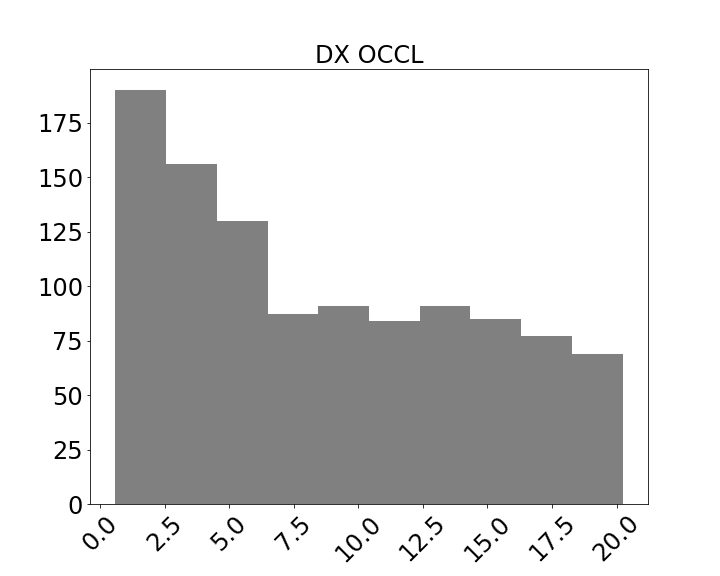}
    \includegraphics[width=0.24\linewidth]{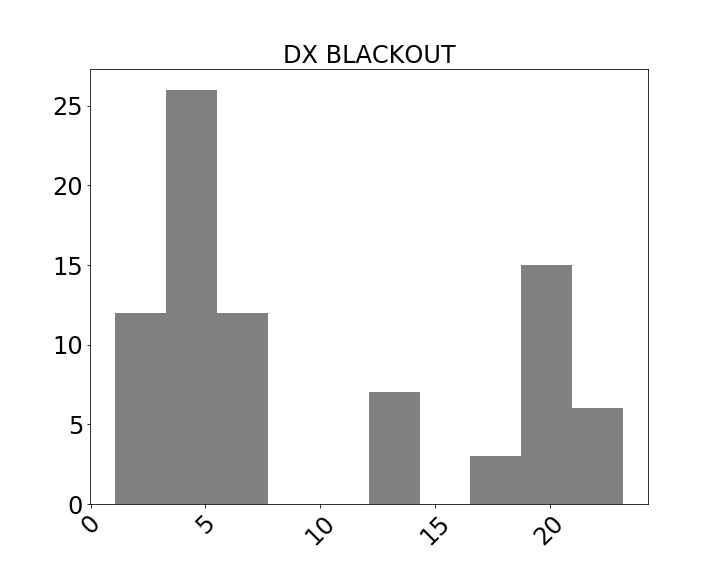}
    \includegraphics[width=0.24\linewidth]{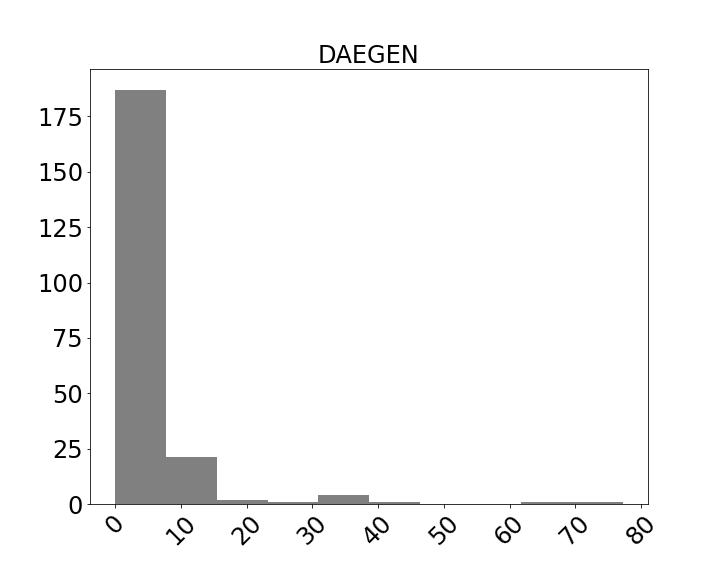}
    \caption{Driving}
  \end{subfigure}
    \caption{Time distributions per dataset.}
    \label{fig:timedistributionsWB}
\end{figure*}

\textbf{Efficiency.}~Figure~\ref{fig:timedistributionsWB} shows the
distributions of time (as histograms) associated with the execution of
each technique on each dataset. As expected for a black-box technique,
\tname{} is significantly more efficient than white-box alternatives.

\begin{center}
\begin{tcolorbox}[enhanced,width=5.2in,center upper,drop shadow southwest,sharp corners]
  \emph{Summary}: We compared \tname{} with two other differential
  techniques (\dx\ and \df{}) on two datasets (MNIST and Driving) and
  various \nn\ models.  \tname{} was the only technique to generate
  adversarial examples in all cases analyzed and was the fastest
  technique. The norms of the examples created are relatively low.
\end{tcolorbox}
\end{center}

\subsection{Black-box (Non-differential) Techniques}
\label{eval:blackbox-DAEGEN}
%%%%%%%%%%%%%%%%%%%%%%%%%%%%%%%%
This section elaborates on the comparison of DAEGEN with black-box techniques.

\subsubsection{Comparison Techniques}
% It builds on the assumption of continuous-valued confidence scores.
% The technique can be used in untargeted and targeted attacks. 

We selected two state-of-the-art black-box non-differential techniques for comparison with \tname{}, namely \simba{}~\ourcite{Guo2019Simba} and \tremba{}~\ourcite{Huang2020}. \simba{} searches for adversarial inputs using the \lnorm\ norm as \tname{} does. At each search iteration, \simba{} samples a vector from a predefined orthonormal basis and either adds that vector or subtracts it to the target image. \tremba{} is a technique that relies on the transferability property of neural networks~\ourcite{Dong2019Evading} to perform adversarial attacks. First, it learns a low-dimensional embedding using a pre-trained model. Then, it performs an efficient search within the embedding space to attack an unknown target network. The method is query efficient and achieves high success rates, thereby producing adversarial perturbations with high-level semantic transferable patterns. The tools implementing these techniques are available from open-source repositories. We used the most recent version of the code and only updated versions of libraries (\eg{}, PyTorch) to assure that the techniques perform at their best.

%Table~\ref{modelxmethodxtechniques_black} shows the \numTechsBB{} black-box non-differential techniques we used to compare against \tname{}. These techniques have been recently used in the evaluation of related techniques. All these techniques are available from public source repositories. We used the most recent version of the code and only updated versions of libraries (\eg{}, Keras/Tensorflow and PyTorch) to make sure these techniques perform at their best.

%%%%%%%%%%%%%%%%%%%%%%%%%%%%%%%%
\subsubsection{Datasets and Models}
%%%%%%%%%%%%%%%%%%%%%%%%%%%%%%%%

We used the \inet{}~\ourcite{Deng2009} dataset in the
evaluation. \inet{} is a large image dataset, containing nearly $15$
million images divided into $22$K classes. As the original dataset is
no longer publicly available, we used the dataset made available by
Huang \etal{} in the evaluation of \tremba{}~\footnote{Available at
  \trembarep{}}~\ourcite{Huang2020}. This reduced dataset contains
$22$K images, with one image per class. Table
\ref{tab:modelxmethodxBBtechniques} shows the datasets and models used
in the evaluation.

\begin{table}[!h]
  \small
  \caption{\label{tab:modelxmethodxBBtechniques}Benchmarks and Models
    used for comparison with Black-Box techniques.}
  \vspace{-1ex}
  \begin{tabular}{ccc}
    \toprule
    \multicolumn{1}{c}{Datasets} & Model (Source) & Used at \\
    \midrule
     \multirow{5}{*}{ImageNet} 
     & Densenet121~\ourcite{Huang2016DenselyCC} & \multirow{4}{*}{\citet{Huang2020}} \\ 
    & MobilenetV2~\ourcite{Sandler2018MobileNetV2IR} & \\
     & ResNet34~\ourcite{He2015DeepRL} & \\
     & VGG16~\ourcite{Simonyan2014deep} & \\
    & VGG19~\ourcite{Simonyan2014deep} & \\
    \bottomrule
  \end{tabular}
\end{table}

In contrast with the comparison against white-box techniques, this experiment only uses one dataset. The rationale is that we found that the implementation of \tremba{} and \simba{} depend on the format of files they accept on input. For example, \simba{} performs algebraic operations on 3-dimensional matrices, representing the images. These operations are invalid on 2-dimensional matrices, such as those from the MNIST dataset. In principle, changing the implementation of these tools to consider different formats is possible. However, we preferred to discard additional datasets instead of changing the code of the tools, as the later could result in the introduction of bias in the evaluation. We used models commonly used to evaluate the performance of input generation techniques on this dataset.

%%%%%%%%%%%%%%%%%%%%%%%%%%%%%%%%
\subsubsection{Setup} 
%%%%%%%%%%%%%%%%%%%%%%%%%%%%%%%%
Every technique uses a budget of $10$K queries to search for an adversarial input, stops when it finds the first adversarial input, and runs on the same set of $100$ randomly selected images from the \inet{} dataset. We used the same number of queries used in prior work and chose 100 as the number of seed images to finish the execution of experiments in an acceptable time. Considering this setup, our script runs the experiments in approximately 21h --\tname{} runs for 10h08m, \simba\ runs for 7h22m, and \tremba\ runs for 3h22m. We ran the techniques according to the guidelines for replicating their experimental results~\ourcite{simbarepository,trembarepository} and calculated DSR according to the definition from Section~\ref{sec:metrics}. Recall that to compare \tname\ (differential) with \simba\ and \tremba\ (non-differential), we created a pipeline of two \nn{}s to simulate differential behavior for the non-differential techniques.

\subsubsection{Results}
Table~\ref{tab:results-black-box} summarizes results. The DSR of
\simba{} and \tremba{} are similar and considerably lower than that of
\tname{}, which was successful in generating adversarial inputs in
99.2\% of the cases. \simba\ produced images with the lowest
\lnorm\ among the techniques and \tremba\ produced images with the
highest \lnorm, almost twice as high as for the images that
\tname\ produced. Considering the number of queries to the models,
\tname{} required a higher number of queries but that number is well
below the 10K budget.

\begin{table}[h!]
  \small
  \caption{\label{tab:results-black-box}Average results comparing \tname{} and Black-Box
    techniques:~\simba{}~\ourcite{Guo2019Simba} and \tremba{}~\ourcite{Huang2020}}
\setlength{\tabcolsep}{10pt}  
\begin{tabular}{ccrrr}
  \toprule
  Dataset & Technique & DSR & \multicolumn{1}{c}{\lnorm{}} & \#Queries\\
  \midrule
%%  & DeepXplore light & 0.732 & 14.034 & 5.573\\
  \multirow{3}{*}{MNIST} & \simba{} & 0.731 & \cellcolor{lightgray}3.147 & 559.0 \\ %38
  & \tremba{} & 0.769 & 11.221 & \cellcolor{lightgray}424.1 \\  %65
  & \tname{} & \cellcolor{lightgray}0.992 & 5.900 & 949.8 \\ %19
  \bottomrule
\end{tabular}
\end{table}

\textbf{Effectiveness.} Recall that \tname{} is a differential
technique. It produces DIAEs (see Section~\ref{sec:introduction}) on
output. Existing black-box techniques are non-differential; they do
\emph{not} produce DIAEs on output. DSR enables us to compare
effectiveness of these techniques on the same grounds and,
consequently, circumvent this problem. Section~\ref{sec:metrics}
defines DSR for differential and non-differential techniques. In both
cases, DSR evaluates effectiveness of a technique \emph{on a pair of
  models}. Non-differential DSR checks, for a given pair of models
$\langle{}a,b\rangle{}$, whether or not the adversarial input produced
by a (non-differential) technique for the model $a$ results in a
different output when fed to $b$. Effectively, DSR measures the
ability of each technique to produce DIAEs.

Table~\ref{tab:results-blackbox} shows DSR values obtained by each
technique for every pair of model combination. The dashes on the
\tname{} section of the table indicate that \tname{} produces the same
results regardless of the order of input models. For example, the
result of \tname{} on the pair
$\langle{}$Densenet121,~MobilenetV2$\rangle$ is the same as
$\langle{}$MobilenetV2,~Densenet121$\rangle$. As results show,
considering the ability to generate DIAEs, \tname{} is superior to
\tremba\ and \simba\ in every one of the 20 combinations of models we
analyzed. In 10 combinations \tname{} had a perfect DSR score. In the
rest of the combinations, \tname{} obtained a score of no less than
97.7\%. In contrast, \tremba{} produced a DSR score above 90\% in
only one of the combinations:~$\langle{}$VGG16,~MobilenetV2$\rangle$.

\begin{table}[!h]
  \small
  \caption{\label{tab:results-blackbox}DSR values for black-box
    techniques for every pair of model combination.}
  \vspace{-1ex}
  \begin{tabular}{ccccccc}
    \toprule
    \multirow{1}{*}{}            & Networks      &  Densenet121    &
    MobilenetV2 &  ResNet34 & VGG16 & VGG19\\
    \midrule                              
    \multirow{5}{*}{\simba{}}   & Densenet121 &        NA &         0.769 &     0.733 &  0.765 &  0.765 \\ 
                             & MobilenetV2 &        0.822 &         NA &     0.842 &  0.839 &  0.835 \\ 
                             & ResNet34    &        0.719 &         0.696 &     NA &  0.711 &  0.700 \\  
                             & VGG16       &        0.744 &         0.700 &     0.728 &  NA & 0.731\\
                             & VGG19       &        0.649 &         0.628 &     0.638 &  0.613 & NA \\
                             
    \midrule 
    \multirow{5}{*}{\tremba{}}   & Densenet121   &  NA   & 0.817 &  0.779 &  0.820 &  0.820\\
                                 & MobilenetV2   &  0.792 & NA   &  0.828 &  0.883 &  0.883  \\ 
                                 & ResNet34      &  0.766 & 0.815 &  NA   &  0.712 &  0.661     \\ 
                                 & VGG16         &  0.807 & 0.909 &  0.767 &  NA   &  0.486    \\
                                 & VGG19         & 0.807  & 0.883 &  0.704 &  0.445 &  NA     \\ 
                                 
\midrule
\multirow{5}{*}{\tname{}}   & Densenet121 & NA           &    1.0   &  1.0    &  1.0 & 1.0\\  
                            & MobilenetV2 & ---          &    NA     &  1.0    &  0.989  & 0.989\\ 
                            & ResNet34    & ---          &    ---     &  NA      &  0.988  &  0.988\\ 
                            & VGG16       & ---          &    ---     &  ---     &  NA    & 0.977 \\
                            & VGG19       & ---          &    ---     &  ---     &  ---   &  NA\\
    % \midrule
    % \multirow{5}{*}{\simba{}} & vgg16 &  -- &  0.731 &  0.728 &     0.744 &      0.700000 \\
    % & densenet121  &  0.764 &  0.764 &  0.732 &     -- &      0.769 \\
    % & mobilenet\_v2 &  0.835 &  0.838 &  0.842 &     0.822 &      --\\
    % & resnet34     &  0.700 &  0.711 &  ---&     0.718 &      0.695 \\
    % & vgg19        &  0.612 &  -- &  0.638 &     0.649 &      0.627 \\

    % \midrule
    % \multirow{5}{*}{\tname{}} & 
    % vgg16          &  -- &  0.977 &  0.988 &     1.00 &    0.988\\
    % & densenet121  &  1.0   &  1.0   &  1.0   &     -- &    0.988\\
    % & mobilenet\_v2 &  0.988 & -- &  0.988 &     0.988 &    -- \\
    % & resnet34     &  0.988 &  0.988 &  -- &     1.00 &    1.00 \\
    % & vgg19        &  0.977 &  -- &  0.988 &     1.00 &    0.988 \\
    \bottomrule
  \end{tabular}
\end{table}

\begin{figure}[h!]
  \centering
  \includegraphics[width=0.28\linewidth]{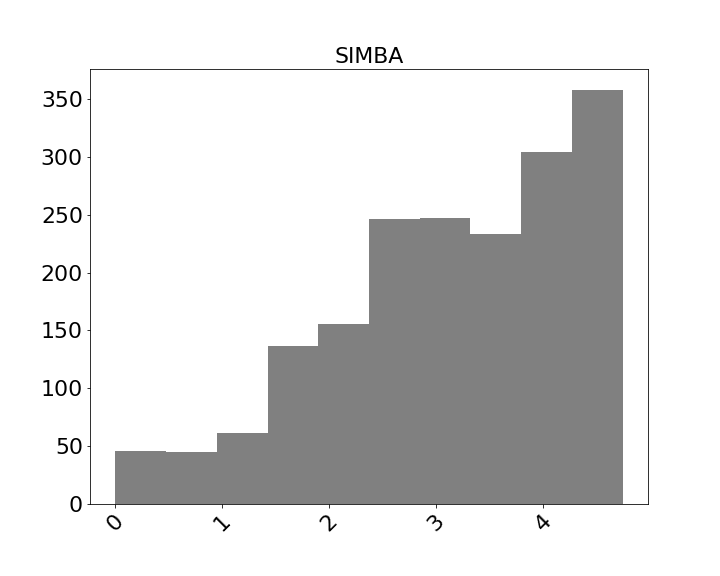}  
  \includegraphics[width=0.28\linewidth]{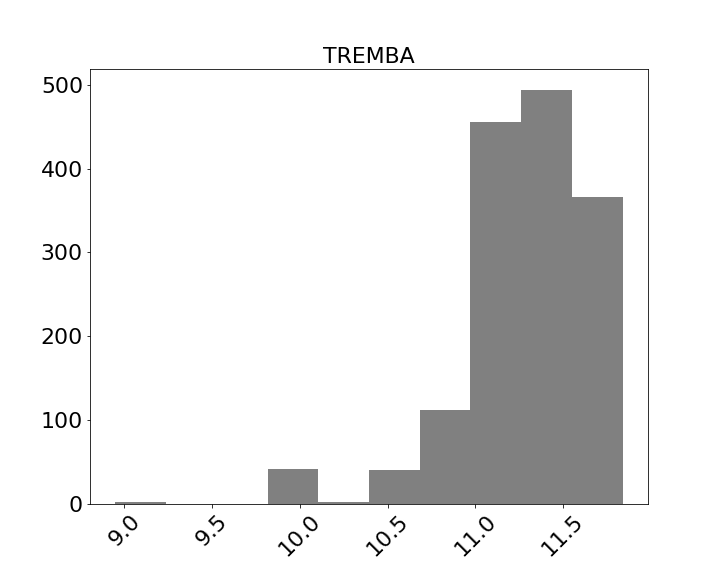}
  \includegraphics[width=0.28\linewidth]{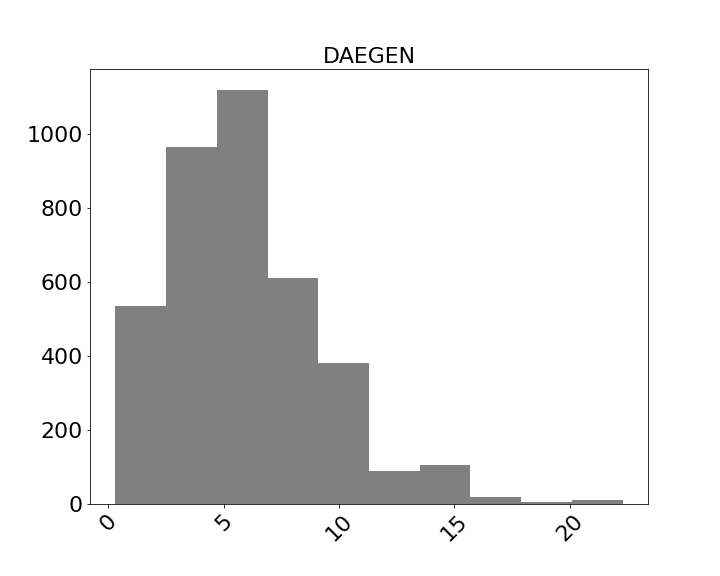}    
  \caption{Distributions of \lnorm.}
  \label{fig:l2disBB}
\end{figure}

\vspace{0.5ex} \textbf{Precision}~Figure~\ref{fig:l2disBB} shows the
distributions of the \lnorm\ norm for each technique. Note that
\simba\ produces examples with the lowest values of \lnorm, but it
produces less examples compared to \tname{}.

\begin{figure}[h!]
    \centering
    \includegraphics[width=0.28\linewidth]{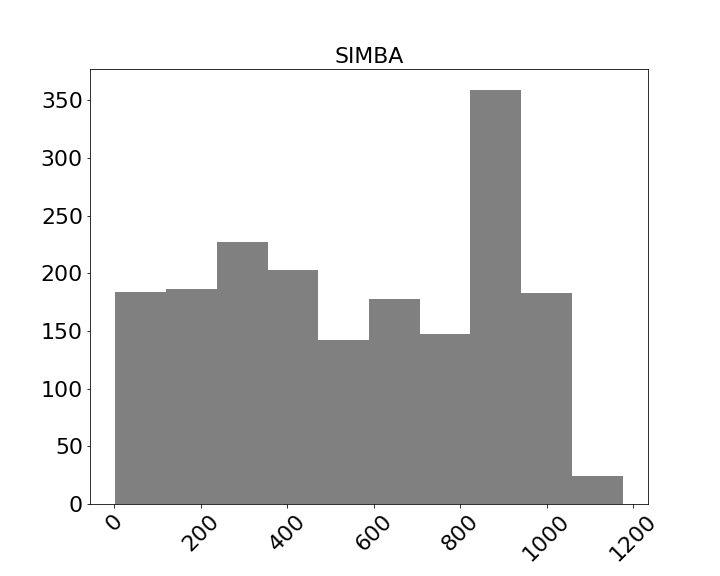}    
    \includegraphics[width=0.28\linewidth]{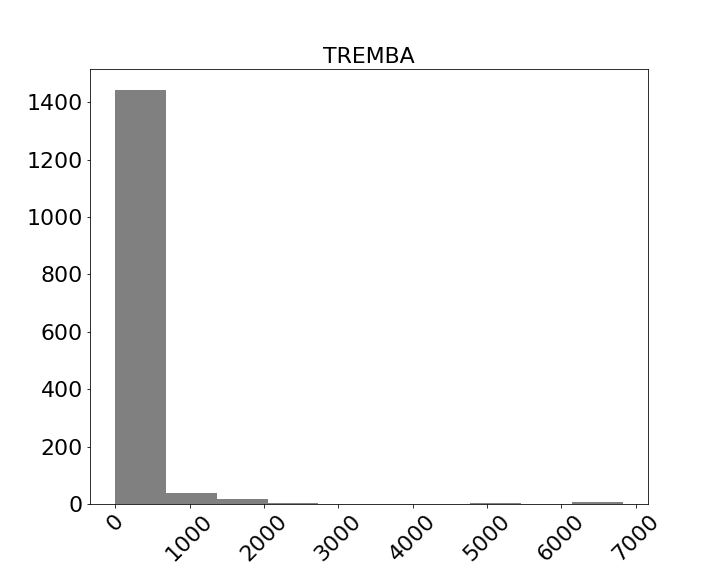}
    \includegraphics[width=0.28\linewidth]{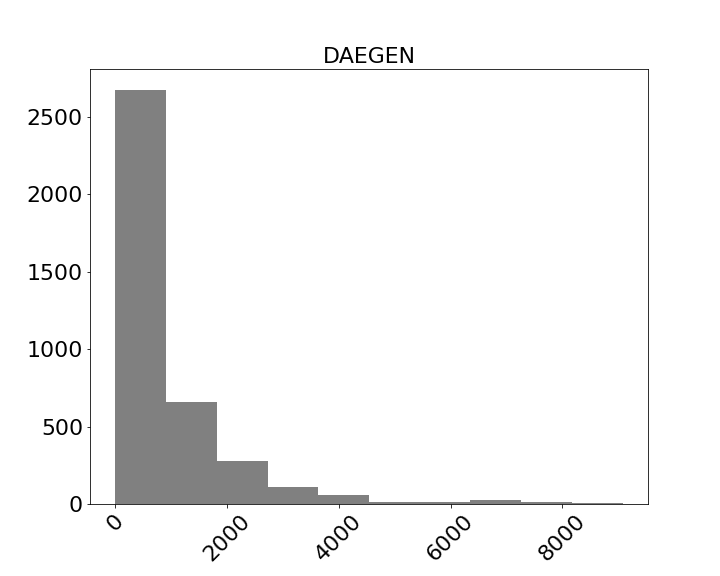}
    \caption{Distributions of number of queries.}
    \label{fig:numqueriesDist}
\end{figure}

\vspace{0.5ex} \textbf{Efficiency.}~Figure~\ref{fig:numqueriesDist}
shows the distributions of number of queries for each
technique. Results indicate that \tname{} required a higher number of
queries compared to alternative techniques.

\begin{center}
\begin{tcolorbox}[enhanced,width=5.2in,center upper,drop shadow southwest,sharp corners]
  \emph{Summary}: We compared \tname{} with two other black-box
  techniques (\simba\ and \tremba) on their ability to produce
  DIAEs. We used a popular dataset (\inet) and five \nn\ models in the
  comparison. \tname{} was able to produce DIAEs in 99.2\% of the
  cases whereas \simba\ and \tremba\ produced DIAEs on rates of 73.1\%
  and 76.9\%, respectively. Considering the norm, \tname{} produced
  images with much lower \lnorm{} values than \tremba\ and slightly
  higher values than \simba{}.
\end{tcolorbox}
\end{center}

\subsection{Threats to Validity}
%-------------------------------------------------------

As is the case of most empirical evaluations, there are external,
internal, and construct threats to the validity of the experimental
results we reported.

\vspace{1ex}\noindent\textbf{External.}~As usual, the extent of
generalization of results are limited to the observations we obtained
from our data, including the datasets, models, and techniques we
selected in this study. To mitigate this threat, we carefully selected
a wide variety of datasets and models available for the corresponding
task. We analyzed the MNIST, \inet, and Driving datasets, which are
popular in this domain. The models we selected are commonly used to
evaluate related techniques, use different architectures, and were
pretrained on the corresponding datasets. Note that there is an
increasing variety of \nn\ architectures available in the
literature. We focused on models previously trained and used for
evaluating related techniques. Considering the techniques, we selected
the state-of-the-art in research. We focused on differential white-box
techniques (namely, \dx{}~\ourcite{pei-etal-sosp2017} and
\df~\ourcite{guo-etal-fse2018}) as to evaluate the impact of the
black-box contribution and on non-differential black-box techniques
(namely, \simba~\cite{Guo2019Simba} and \tremba~\ourcite{Huang2020})
as to evaluate the impact of the differential contribution.

\vspace{1ex}\noindent\textbf{Internal.}~As typical for a human
activity (coding), the implementation of \tname{} or the scripts we
used to evaluate \tname\ could have (unintentional) problems. To
mitigate this issue, the authors cross-checked the results reported in
this section. Every author analyzed the data individually and pointed
to incoherences when they appeared.

\vspace{1ex}\noindent\textbf{Construct.}~As usual, the plausibility of
the metrics we used to evaluate \tname\ could be questioned. We
evaluated \tname\ on effectiveness, precision, and efficiency as we
consider these key quality indicators of an input generation
technique. Considering effectiveness, we measured the ability of
techniques to produce DIAEs, which is the central goal of our
technique. To fairly compare non-differential with differential
techniques, we needed to simulate differential behavior, as described
on Section~\ref{sec:eval-effectiveness}. We made our best to fairly
evaluate black-box (non-differential) techniques and we believe the
evaluation is valuable, but these techniques were not designed for
that purpose. Considering precision, there are different norms in the
literature to evaluate image similarity. The Euclidean distance
(\lnorm) is the most popular norm in related research.

\section{Related Work}
\label{sec:related}
%------------------------------------------------------------

%\Mar{Moved from intro to here --->}

%% We discuss related work from the perspectives of the challenges due to
%% the black-box setting, existing work on differential testing on neural
%% networks, and optimization-based adversarial example generation
%% techniques.

This section discusses most related work.

%%%%%%%%%%%%%%%%%%%%%%%%%%%%
\subsection{Black-box Input Generation}
\label{sec:white-and-black-box}
%%%%%%%%%%%%%%%%%%%%%%%%%%%%
%White-box and Black-box Input Generation.}
\sloppy

There are three main categories of adversarial attacks: white-box,
gray-box, and
black-box~\ourcite{Serban2018AdversarialE}~\ourcite{xu2019adversarial}.
Many white-box techniques have been proposed recently in the
literature. These technique typically use a gradient search to
generate adversarial examples (e.g., Fast Gradient Sign Method
(FGSM)~\ourcite{Goodfellow2014}, Projected Gradient Descent
(PGD)~\ourcite{Carlini2016TowardsET}~\ourcite{Madry2018towards}, and
$L_\infty$ Basic Iterative Attack~\ourcite{Kurakin2017}). See
\citet{Yuan2017AdversarialEA} and \citet{Zhang2019Defending} for a
survey of techniques.
% xiaowei: don't need an introduction of gradient descent in related works
%Gradient-descent is an iterative optimization algorithm used to minimize some function, where on each step, it moves to the negative direction of the gradient, which corresponds to the maximum slope direction (i.e., the steepest descent)~\ourcite{bertsekas1997nonlinear}~\ourcite{Luenberger2015}. 
A white-box attack is better suited when the attacker has access to the implementation of the \nn{}, 
%i.e., the attacker has access to the gradients of the \nn{}, 
which is more likely to occur during the development and testing
phases.  Once the model is deployed, it can be seen as a black-box
system~\ourcite{Yuan2017AdversarialEA}, making it hardly practical to
apply white-box approaches. It is also worth noting that techniques
have been recently proposed to defend against gradient-based
attacks. Such defenses can manipulate the gradients performing
gradient masking or
obfuscation~\ourcite{Alzantot2019}~\ourcite{Athalye2018ObfuscatedGG}.
In a gray-box setting, the attacker has some knowledge about the
target network (e.g., architecture, train dataset, or defenses
mechanism), but not the gradients.  In a black-box setting, the target
model is available as a black-box function. Therefore, the attacker is
oblivious to critical and essential information about the target
model. This is a realistic characterization of the availability of NN
models in the real-world.  \ourcite{Serban2018AdversarialE} describe
adversarial examples and a threat model for this kind of attack.

%From the listed ways of attacks, 13\% were classified as black-box attacks.

Most existing approaches of black-box attacks are considered ineffective since they require hundreds of thousands of queries to the model to generate a single adversarial example~\ourcite{Chen2017Zoo}~\ourcite{Mosli2019}. 
Depending on the model, this could be time-consuming and rely on substantial computational power, therefore, making the attack not practical.
Another disadvantage is that some methods often propose the design and implementation of a surrogate model to approximate the decision boundary of the target \nn{} and then apply gradient-descent optimization on the surrogate model, which is also the case in white-box settings~\ourcite{Huang2020}. 
However, in a black-box setting, the limited number of queries makes these approaches hard to apply as the surrogate model has to, in broad terms, learn the outputs of the target model.
Also, training a surrogate model can be challenging due to the complexity and non-linearity in the classification rules, which can degrade the performance of black-box methods~\ourcite{Tu2018AutoZOOMAZ}.
Also, it has been found that adversarial training (i.e., using adversarial examples to retrain the model~\ourcite{Ilyas2019AdversarialEA}) combined with gradient-descent techniques (e.g., FGSM and PGD) makes \nn{} models more robust to white-box attacks~\ourcite{Tramer2018ensemble}. By inference, this combination can be the case of surrogate-based black-box attacks.
The implication of this is that models submitted to adversarial training may be robust to surrogate techniques. However, it can still be affected by adversarial examples generated by black-box techniques~\ourcite{Tramer2018ensemble}, e.g., those that apply local or global search meta-heuristics to generate adversarial examples.

% Also, it has been found that adversarial training (i.e., using adversarial examples to retain the model~\ourcite{Ilyas2019AdversarialEA}) combined with gradient-descent techniques (e.g., FGSM and PGD),  are more robust to white-box than to black-box attacks due to gradient masking~\ourcite{Tramer2018ensemble}. The implication of this is that models submitted to adversarial training are robust to white-box techniques but can still be affected by adversarial examples generated by black-box techniques. 

%\subsection{\Mar{Revise...}Other Stuff}

%%%%%%%%%%%%%%%%%%%%%%%%%%%%
\subsection{Differential Testing}
%%%%%%%%%%%%%%%%%%%%%%%%%%%%

%\noindent\textbf{Differential testing}. 
\citet{Pei2019} proposed \dx, a white-box testing framework for
large-scale deep learning systems. \dx{} formulates search as a joint
optimization problem of generating inputs that maximize neuron
coverage and maximize the differential behaviors (or merely the
differences) of multiple similar DL systems. Besides the white
vs. black box differences, DeepXplore and \tname{} formulate different
optimization problems. DeepXplore maximizes differential behavior
across various neural nets, while \tname{} considers only two neural
networks during the attack.

%\noindent\textbf{Adversarial black-box attacks}.

%\ourcite{Papernot2017} 
%demonstrate that black-box attacks were a practical option for real-world adversaries without any knowledge about the
%models they were attacking. Their work provides the first demonstration of a DNN model being remotely controlled by an attacker
%in a black-box setting. Their attack strategy is to train a substitute
%DNN with a synthetic dataset to approximate the decision boundaries of the target DNN: the inputs are synthetic and generated by
%the adversary. At the same time, the outputs are labels assigned
%by the target DNN and observed by the adversary. Adversarial
%examples are crafted using the substitute parameters, e.g., by applying gradient-descent optimization proposed by \ourciteauthor{Goodfellow2014}
%and \ourciteauthor{Papernot2015TheLO}. \tname{}, on the other hand, applies gradient free optimization to craft adversarial examples, meaning it doesn’t
%need to access the parameters (weights) of any models. Although it
%doesn’t require a substitute model to learn the outputs of a target
%model, it does rely on the output of two models, to proceed with
%the optimization.

The solution proposed by \citet{Papernot2017} also involves multiple
models. Their solution builds on the concept of transfer-based
attacks, i.e., the adversarial examples crafted for the surrogate
models can be then used to attack other models. The adversarial
examples prepared by \tname{} were not tested against other models
besides the models involved in the attack. We can only say that, in
some cases, it can attack both models in the attack configuration, but
in the other cases, it can only attack one of the models.

%%%%%%%%%%%%%%%%%%%%%%%%%%%%%%%%%%%%%
\subsection{Optimization-based Adversarial Example Generation}
%%%%%%%%%%%%%%%%%%%%%%%%%%%%%%%%%%%%%

In recent studies, we note the use of optimization techniques to generate adversary examples. In these studies, there exists some concern in minimizing the amount of queries needed to rediscover the attack until an adverse example is produced. ~\ourcite{ Chen2019} and \ourcite{Alzantot2019} have proposed genetic algorithms. In the first work, a new optimized black-box attack was proposed, based on the genetic algorithm (POBA-GA), capable of generating approximate optimal opponents through evolutionary operations, including initialization, selection, crossover, and mutation.  In the second work, a new gradient-free optimization technique (GenAttack) that uses genetic algorithms to generate adversary examples was proposed. GenAttack can successfully attack some advanced defense techniques for ImageNet, suggesting that evolutionary algorithms are promising in promoting black-box attacks. Both works perform targeted attacks (while \tname{} only performs non-targeted attacks) and formulate their optimization problem as a multi-objective optimization problem, where they try to maximize the probability P(y|X) for a specified y. ~\ourcite{Chen2019} also try to minimize the difference between the original image and the adversarial image, and \ourcite{Alzantot2019} try to minimize the probability of $P(\neg y|X)$. Therefore, the difference between theses threat models and \tname{} relies on the fact that it solves the optimization problem by trying to maximize the difference between the probability $P(y|X)_a$ and the probability $P(y|X)_b$, for a pair of models $M_a$ and $M_b$ respectively.

Other more sophisticated techniques such as Zeroth Order Optimization have been applied in \ourcite{Chen2017Zoo} and \ourcite{Cheng2019}. Particle swarm optimization (PSO) is present in the works \ourcite{Mosli2019} and \ourcite{Zhang2019}. PSO can quickly converge on sufficiently good solutions with few queries. Like the Genetic Algorithm, PSO also requires the design of fitness function. Both \ourcite{Mosli2019} and \ourcite{Zhang2019} opt for fitness functions that are equations containing the output of the network for a given particle (possible adversarial solution) and the L2 norm of the original image and the particle.

Although many meta-heuristics optimization approaches have natural methods to randomly introduce small noises and perturbations into inoffensive image samples in their search for feasible (or optimal) solutions, we must mention that they are not the only options. Generative Adversarial Networks (GANs) \ourcite{Goodfellow2014GenerativeAN}  are also viable options to generate candidate solutions for the optimization algorithms~\ourcite{Xiao2018GeneratingAE}~\ourcite{Fang2019}. Bayesian optimization~\ourcite{Shukla2019} and random fuzzing~\ourcite{Zhang2019} are also interesting approaches for black-box attacks.  However, these techniques are usually a more sophisticated solution for the adversarial generation problem. For the sake of simplicity, we chose to apply custom implementations of search-based optimization techniques, namely, Hill Climbing,  with minimal modifications to satisfy our attack-method requirements.

\section{\uppercase{Conclusions}}
\label{sec:conclusion}
%------------------------------------------------------------

We presented \tname{}, a black-box method to exploit differential
behavior between two neural networks by finding difference-inducing
adversarial examples (DIAEs) that maximize their output difference
(under constraints on the perturbation). We conducted a comprehensive
set of experiments to evaluate \tname. We considered existing white-box methods that can generate DIAEs and then adapted existing black-box methods to work on DIAEs. Our experiments show the superiority of \tname{} over all of them in terms of effectiveness, precision, and efficiency. 

We compared \tname{} with two other differential techniques (\dx\ and
\df{}) on two datasets (MNIST and Driving) and various
\nn\ models. \tname{} was the only technique to generate adversarial
examples in all cases analyzed and was the fastest technique. Besides,
we compared \tname{} with two other non-differential black-box
techniques (\simba\ and \tremba) on their ability to produce DIAEs. We
used a popular dataset (\inet) and five \nn\ models in the
comparison. We found that \tname{} was able to produce DIAEs in $99.2$\% of the cases, whereas \simba\ and \tremba\ produced DIAEs on rates of 73.1\% and 76.9\%, respectively.  

Our findings can help in the development of techniques to test,
evaluate, and ensure robustness of neural networks against black-box
adversarial attacks, especially in security-critical domains.
%given the similarity of \tname{} with evolutionary black-box
%fuzzing.
Future work will focus on improving DSR to achieve a $100$\% success
rate and validating our approach against known defense methods. It will also be a natural next step to see whether \tname{} can be used to support explainable AI. We believe that a good explanation of neural network behavior may be achievable through identifying the differences between neural networks. 

\balance
\bibliographystyle{ACM-Reference-Format}
%\setcitestyle{authoryear, open={((},close={))}
\bibliography{main}
%\section*{\uppercase{Appendix}}

% \noindent If any, the appendix should appear directly after the
% references without numbering, and not on a new page. To do so please use the following command:
% \textit{$\backslash$section*\{APPENDIX\}}

\end{document}